\pdfoutput=1
\documentclass{article}

\PassOptionsToPackage{table}{xcolor}
\usepackage{arxiv}

\usepackage[utf8]{inputenc}
\usepackage[T1]{fontenc}
\usepackage{microtype}
\usepackage{times}          
\usepackage{latexsym}
\usepackage{xcolor}

\usepackage{amsmath,amssymb,mathtools}

\usepackage{booktabs}
\usepackage{graphicx}
\usepackage{multirow}
\usepackage{multicol}
\usepackage{makecell}
\usepackage{array}
\usepackage{tabularx}
\usepackage{longtable}
\usepackage{rotating}
\usepackage{pdflscape}
\usepackage{colortbl}
\usepackage{graphicx}
\usepackage{caption}
\usepackage{soul}
\usepackage{subcaption}
\usepackage{float}
\usepackage{wrapfig}
\usepackage[table]{xcolor}
\usepackage{colortbl}
\usepackage{array}
\usepackage{booktabs}
\usepackage{makecell}
\usepackage{adjustbox}

\usepackage{xcolor}

\usepackage{enumitem}
\usepackage{algorithm}
\usepackage{algorithmic}

\usepackage{listings}

\usepackage{tikz}
\tikzset{repoitem/.style={rectangle, draw, fill=blue!10, rounded corners, minimum height=0.8cm, minimum width=2cm, align=center}}

\usepackage{pgfplots}
\usetikzlibrary{shapes,arrows,positioning,calc}
\pgfplotsset{compat=1.18}
\usepackage{natbib}
\renewcommand\cite{\citep}  

\usepackage{hyperref}
\usepackage{url}
\usepackage{doi}

\definecolor{niceblue}{RGB}{0,102,204}

\hypersetup{
    colorlinks=true,
    linkcolor=niceblue,
    citecolor=niceblue,
    urlcolor=niceblue
}

\usepackage{xcolor}
\usepackage{tikz}
\usetikzlibrary{arrows.meta, positioning}
\usepackage[most]{tcolorbox}
\usepackage{pgfplots}
\pgfplotsset{compat=1.18}

\definecolor{GNNpurple}{RGB}{103,58,183}
\definecolor{GNNblue}{RGB}{21,101,192}
\definecolor{GNNteal}{RGB}{0,137,123}
\definecolor{GNNgreen}{RGB}{46,125,50}
\definecolor{GNNorange}{RGB}{230,81,0}
\definecolor{GNNred}{RGB}{183,28,28}
\definecolor{GNNgold}{RGB}{255,193,7}
\definecolor{GNNpink}{RGB}{173,20,87}
\definecolor{GNNgray}{RGB}{66,66,66}
\definecolor{GNNcyan}{RGB}{0,131,143}
\definecolor{catHomo}    {RGB}{21,101,192}   
\definecolor{catHetero}  {RGB}{0,150,136}    
\definecolor{catTemporal}{RGB}{230,81,0}     
\definecolor{catSpatial} {RGB}{121,85,72}    
\definecolor{catHier}    {RGB}{103,58,183}   
\definecolor{catMol}     {RGB}{100,125,50}    
\definecolor{catGen}     {RGB}{173,20,87}    
\definecolor{diffEasy}{RGB}{76,175,80}
\definecolor{diffMedium}{RGB}{255,152,0}
\definecolor{diffHard}{RGB}{244,67,54}
\definecolor{diffExpert}{RGB}{156,39,176}
\colorlet{catHeader}{GNNblue!85}
\definecolor{placeholder}{RGB}{128,128,128}
\definecolor{GNNpurple}{RGB}{103,58,183}
\definecolor{GNNblue}  {RGB}{21,101,192}
\definecolor{GNNteal}  {RGB}{0,137,123}
\definecolor{GNNgreen} {RGB}{46,125,50}
\definecolor{GNNorange}{RGB}{230,81,0}
\definecolor{GNNred}   {RGB}{183,28,28}
\definecolor{GNNpink}  {RGB}{173,20,87}
\definecolor{GNNgray}  {RGB}{66,66,66}
\definecolor{GNNcyan}  {RGB}{0,131,143}
\definecolor{GNNamber} {RGB}{255,152,0}

\newcommand{\catbox}[2]{\colorbox{#1}{\textcolor{white}{\small\textbf{#2}}}}
\newcommand{\diffbox}[2]{\colorbox{#1}{\textcolor{white}{\footnotesize\textbf{#2}}}}

\title{GNN-CB: A Graph Neural Network Competition Benchmark for Human and LLM Evaluation}
\author{%
\begin{minipage}{\textwidth}
\centering
\normalfont\normalsize
Murad~Hossen\textsuperscript{1}\thanks{\ \raggedright\hangindent=1.2em\hangafter=1 Corresponding authors:
\protect\nolinkurl{mhossen2@cougarnet.uh.edu},
\protect\nolinkurl{i.rekik@imperial.ac.uk}; \url{https://basira-lab.com/}}\thanks{\ \raggedright Co-first authors: These authors contributed equally.},
Tasneem~Selim\textsuperscript{2}\footnotemark[2],
Gurur~Gamgam\textsuperscript{3}\footnotemark[2],
Tuga~Yousif\textsuperscript{4}\footnotemark[2],
Abderrahmane~Kasmi\textsuperscript{5}\footnotemark[2],
Ikram~Aissiou\textsuperscript{6},
Mubaraq~Onipede\textsuperscript{7},
Faran~Taimoor~Butt\textsuperscript{8},
Sanae~Zrigui\textsuperscript{9},
Rosa~Y.~G.~Paccotacya-Yanque\textsuperscript{10},
Ignatius~Balayo\textsuperscript{11},
Ikram~Elhouiti\textsuperscript{12},
Hadil~Affes\textsuperscript{13},
Bijay~Adhikari\textsuperscript{14},
Sargam~Goyal\textsuperscript{15},
Muhammad~Ibrahim~Isah\textsuperscript{16},
Mohammad~Idrees~Bhat\textsuperscript{17},
Samuel~Kangoni~Matia\textsuperscript{18},
Peguy~Kem-Meka~Tiotsop~Kadzue\textsuperscript{19,20,21},
Maha~Trabelsi\textsuperscript{13},
Emmanuel~Owusu\textsuperscript{22},
Vinit\textsuperscript{23},
Nour~Majdoub\textsuperscript{24},
Tamiru~Alemnew\textsuperscript{25},
Islem~Rekik\textsuperscript{26}\footnotemark[1]
\\[5pt]
\rule{0.35\textwidth}{0.4pt}
\\[5pt]
\normalfont\footnotesize\linespread{0.97}\selectfont
\textsuperscript{1}University of Houston, Texas,USA\,$\cdot$\,
\textsuperscript{2}Department of Mathematics and Computer Science, Faculty of Science, Alexandria University, Alexandria, Egypt\,$\cdot$\,
\textsuperscript{3}Bogazici University, Turkey\,$\cdot$\,
\textsuperscript{4}Ankara Yıldırım Beyazıt University, Turkey\,$\cdot$\,
\textsuperscript{5}ESI, Algiers, Algeria\,$\cdot$\,
\textsuperscript{6}University of Algiers 1, Algeria\,$\cdot$\,
\textsuperscript{7}York St John University London, UK\,$\cdot$\,
\textsuperscript{8}Air University, Islamabad, Pakistan\,$\cdot$\,
\textsuperscript{9}LSISI, ENSA, Mohammed Premier University, Oujda, Morocco\,$\cdot$\,
\textsuperscript{10}Universidad Católica San Pablo, Arequipa, Peru\,$\cdot$\,
\textsuperscript{11}Busitema University, Uganda\,$\cdot$\,
\textsuperscript{12}University of Laghouat, Algeria\,$\cdot$\,
\textsuperscript{13}ISI, University of Tunis El Manar, Tunisia\,$\cdot$\,
\textsuperscript{14}Tribhuvan University, Nepal\,$\cdot$\,
\textsuperscript{15}IIT Roorkee, India\,$\cdot$\,
\textsuperscript{16}Shobhit Institute of Engineering and Technology, India\,$\cdot$\,
\textsuperscript{17}MIT World Peace University, Pune, India\,$\cdot$\,
\textsuperscript{18}University of Kinshasa, DR Congo\,$\cdot$\,
\textsuperscript{19}University of Bertoua, Cameroon\,$\cdot$\,
\textsuperscript{20}University of the Witwatersrand, Johannesburg, South Africa\,$\cdot$\,
\textsuperscript{21}African Institute for Mathematical Sciences, Research and Innovation Centre (AIMS RIC), Kigali, Rwanda\,$\cdot$\,
\textsuperscript{22}KNUST, Ghana\,$\cdot$\,
\textsuperscript{23}NSUT Delhi, India\,$\cdot$\,
\textsuperscript{24}ISIMM Monastir, Tunisia\,$\cdot$\,
\textsuperscript{25}Addis Ababa University, Ethiopia\,$\cdot$\,
\textsuperscript{26}BASIRA Lab,
Department of Computing, Imperial College London, United Kingdom
\\[5pt]
\normalfont\small\itshape
All authors are co-affiliated with the BASIRA Lab,
Department of Computing, Imperial College London, United Kingdom.
\end{minipage}%
}

\date{}  
\renewcommand{\undertitle}{}   
\renewcommand{\headeright}{}   
\renewcommand{\shorttitle}{\textsc{GNN-CB}: A Graph Neural Network Competition Benchmark}
\hypersetup{
pdftitle={GNN-CB: A Graph Neural Network Competition Benchmark for Human and LLM Evaluation},
pdfauthor={Murad Hossen, Tasneem Selim, Gurur Gamgam, Tuga Yousif, Abderrahmane Kasmi, Ikram Aissiou, Mubaraq Onipede, Faran Taimoor Butt, Sanae Zrigui, Rosa Y. G. Paccotacya-Yanque, Ignatius Balayo, Ikram Elhouiti, Hadil Affes, Bijay Adhikari, Sargam Goyal, Muhammad Ibrahim Isah, Mohammad Idrees Bhat, Samuel Kangoni Matia, Peguy Kem-Meka Tiotsop Kadzue, Maha Trabelsi, Emmanuel Owusu, Vinit, Nour Majdoub, Tamiru Alemnew, Islem Rekik},
}

\usepackage{etoolbox}
\makeatletter
\preto\@maketitle{\setlength{\tabcolsep}{0pt}%
  \def\@makefnmark{\hbox{\@textsuperscript{\normalfont\@thefnmark}}}}
\makeatother

\colorlet{GNNsky}{black}
\colorlet{GNNlavender}{black}
\colorlet{GNNmint}{black}
\colorlet{GNNcoral}{black}

\begin{document}
\maketitle

\begin{abstract}
Large language models (LLMs) have demonstrated strong performance on coding and reasoning benchmarks; however, their ability to solve graph-structured machine learning problems remains largely unexplored. In particular, no benchmark currently evaluates whether LLMs can autonomously solve end-to-end Graph Neural Network (GNN) coding tasks under realistic competition settings. To address this gap, this paper introduces \textbf{GNN-CB}, the first competition-based benchmark for evaluating both humans and LLMs on GNN coding tasks. GNN-CB consists of 18 curated competitions spanning node-, edge-, and graph-level prediction across diverse graph categories, domains, and difficulty tiers. All submissions are evaluated through a unified automated pipeline with hidden test sets and standardized scoring. Human participants solve tasks under controlled competition constraints, while LLMs are evaluated using a frozen zero-shot prompting protocol based on a plan-then-code paradigm with bounded execute-and-repair loops. The benchmark additionally supports both non-agent and autonomous agent-based evaluation within the same protocol. Under our evaluated protocol, LLMs rarely match Human Top performance and show less stable performance across competitions. No single model dominates: a few competitions are won by LLMs, yet humans still hold the top score on most tasks. We release GNN-CB as a living benchmark with automated evaluation infrastructure, dynamic leaderboards, and reproducible execution pipelines. Beyond benchmarking, GNN-CB provides a practice-oriented resource for studying GNN implementation across progressively diverse graph-learning tasks. The benchmark and evaluation framework are publicly available at \url{https://basiralab.github.io/GNN-CB/}.

\end{abstract}

 \section{Introduction}
 \label{sec:intro}

\tikzset{
  chip/.style 2 args={
    draw=#1!70, fill=#1!15, text=#1!90!black,
    rounded corners=3pt,
    font=\tiny\bfseries,
    inner xsep=4pt, inner ysep=2pt,
    minimum height=9pt
  },
  hdr/.style={
    draw=#1!70, fill=#1!20,
    rounded corners=4pt,
    font=\tiny\bfseries,
    inner xsep=6pt, inner ysep=3pt,
    minimum width=6.8cm, minimum height=9pt,
    text=#1!60!black
  },
  infobox/.style={
    draw=gray!40, fill=gray!6,
    rounded corners=3pt,
    font=\tiny,
    inner xsep=5pt, inner ysep=3pt,
    align=center
  },
  terminal/.style={
    draw=#1!60, fill=#1!20,
    rounded corners=5pt,
    font=\tiny\bfseries,
    inner xsep=8pt, inner ysep=4pt,
    text=#1!60!black, align=center
  },
  section/.style={
    draw=gray!40, dashed,
    rounded corners=6pt,
    fill=none
  },
  arr/.style={->, >=stealth, thick, gray!70}
}

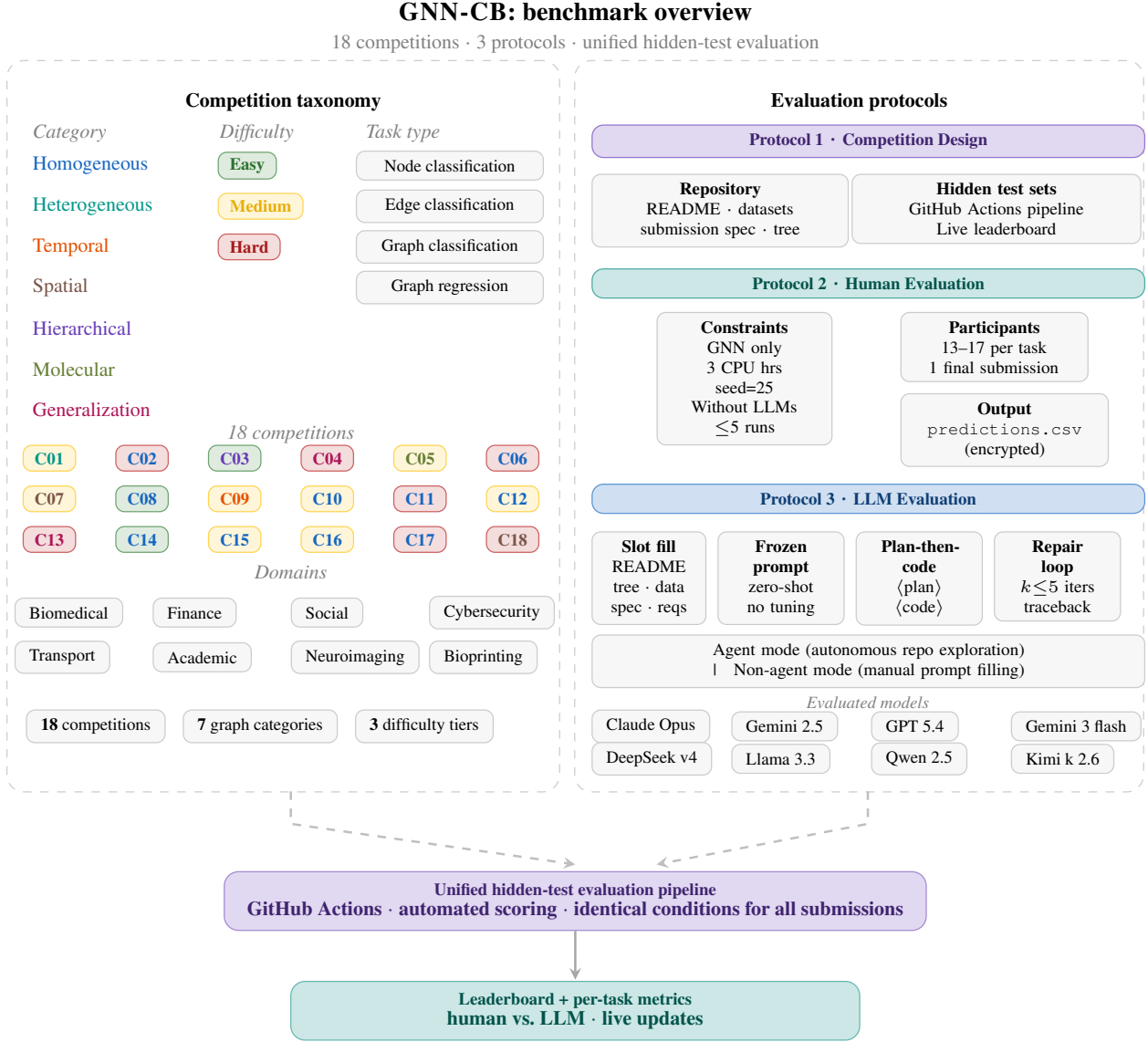
\begin{figure*}[!t]
\centering
\resizebox{\textwidth}{!}{%
\begin{tikzpicture}[
    node distance = 0pt,
    every node/.style = {font=\tiny}
]

\node[font=\small\bfseries] (title) at (7,0)
  {\textsc{GNN-CB}: benchmark overview};
\node[font=\scriptsize, text=gray, below= -2 pt of title] (subtitle)
  {18 competitions $\cdot$ 3 protocols $\cdot$ unified hidden-test evaluation};

\node[section, minimum width=6.8cm, minimum height=9cm,
      anchor=north west] (leftpanel) at (0,-0.6) {};
\node[font=\scriptsize\bfseries, anchor=north] at (leftpanel.north)
  [yshift=-8pt] {Competition taxonomy};

\node[font=\scriptsize, text=gray, anchor=west]
  (cat_lbl) at (0.2, -1.5) {\textit{Category}};

\node[text=catHomo, font=\scriptsize, anchor=west] at (0.2,-1.9) {Homogeneous};
\node[text=catHetero, font=\scriptsize, anchor=west] at (0.2,-2.4) {Heterogeneous};
\node[text=catTemporal,font=\scriptsize, anchor=west] at (0.2,-2.9) {Temporal};
\node[text=catSpatial,font=\scriptsize, anchor=west] at (0.2,-3.4) {Spatial};
\node[text=catHier,   font=\scriptsize, anchor=west] at (0.2,-3.9) {Hierarchical};
\node[text=catMol,    font=\scriptsize, anchor=west] at (0.2,-4.4) {Molecular};
\node[text=catGen,    font=\scriptsize, anchor=west] at (0.2,-4.9) {Generalization};

\node[font=\scriptsize, text=gray, anchor=west]
  at (2.5,-1.5) {\textit{Difficulty}};
\node[chip={GNNgreen}{}, anchor=west]
  at (2.6,-1.9) {Easy};

\node[chip={GNNgold}{}, anchor=west]
  at (2.6,-2.4) {Medium};

\node[chip={GNNred}{}, anchor=west]
  at (2.6,-2.9) {Hard};

\node[font=\scriptsize, text=gray, anchor=west]
  at (4.3,-1.5) {\textit{Task type}};
\node[infobox, anchor=west, minimum width=2.3cm] at (4.3,-1.9) {Node classification};
\node[infobox, anchor=west, minimum width=2.3cm] at (4.3,-2.4) {Edge classification};
\node[infobox, anchor=west, minimum width=2.3cm] at (4.3,-2.9) {Graph classification};
\node[infobox, anchor=west, minimum width=2.3cm] at (4.3,-3.4) {Graph regression};

\node[font=\scriptsize, text=gray] at (3.5,-5.2) {\textit{18 competitions}};


\foreach \id/\textcol/\fillcol [count=\i from 0] in {
    C01/catHetero/GNNgold,
    C02/catHomo/GNNred,
    C03/catHier/GNNgreen,
    C04/catGen/GNNred,
    C05/catMol/GNNgold,
    C06/catHomo/GNNred}{
  \node[
      chip={\fillcol}{},
      text=\textcol,
      anchor=west
  ] at ({0.2 + \i * 1.14}, -5.5) {\id};
}

\foreach \id/\textcol/\fillcol [count=\i from 0] in {
    C07/catSpatial/GNNgold,
    C08/catHomo/GNNgreen,
    C09/catTemporal/GNNgold,
    C10/catHomo/GNNgold,
    C11/catHomo/GNNred,
    C12/catHomo/GNNgold}{
  \node[
      chip={\fillcol}{},
      text=\textcol,
      anchor=west
  ] at ({0.2 + \i * 1.14}, -6) {\id};
}
\foreach \id/\textcol/\fillcol [count=\i from 0] in {
    C13/catGen/GNNred,
    C14/catHomo/GNNgreen,
    C15/catHomo/GNNgold,
    C16/catHomo/GNNgold,
    C17/catHomo/GNNred,
    C18/catSpatial/GNNred}{
  \node[
      chip={\fillcol}{},
      text=\textcol,
      anchor=west
  ] at ({0.2 + \i * 1.14}, -6.5) {\id};
}

\node[font=\scriptsize, text=gray] at (3.5,-6.9) {\textit{Domains}};
\foreach \d [count=\i from 0] in {
    Biomedical, Finance, Social, Cybersecurity,
    Transport, Academic, Neuroimaging, Bioprinting}{
  \node[infobox, font=\tiny, anchor=west]
    at ({0.1 + mod(\i,4) * 1.7}, {-7.4 - floor(\i/4)*0.55})
    {\d};
}

\node[infobox, minimum width=0.8cm] (s1) at (1.1,-8.8)
  {\textbf{18} competitions};
\node[infobox, minimum width=0.8cm, right=6pt of s1] (s2)
  {\textbf{7} graph categories};
\node[infobox, minimum width=0.8cm, right=6pt of s2]
  {\textbf{3} difficulty tiers};

\node[section, minimum width=7cm, minimum height=9cm,
      anchor=north east] (rightpanel) at (14,-0.6) {};
\node[font=\scriptsize\bfseries, anchor=north] at (rightpanel.north)
  [yshift=-8pt] {Evaluation protocols};

\node[hdr={GNNpurple}, anchor=west] (p1hdr) at (7.2,-1.6)
  {Protocol 1 \,·\, Competition Design};

\node[infobox, anchor=north west, minimum width=3.0cm,
      text width=2.8cm] (repo) at (7.2,-2)
  {\textbf{Repository}\\
   README $\cdot$ datasets\\
   submission spec $\cdot$ tree};
\node[infobox, anchor=north west, minimum width=3.4cm,
      text width=3.2cm] (hidden) at (10.4,-2)
  {\textbf{Hidden test sets}\\
   GitHub Actions pipeline\\
   Live leaderboard};

\node[hdr={GNNteal}, anchor=west] (p2hdr) at (7.2,-3.35)
  {Protocol 2 \,·\, Human Evaluation};

\node[infobox, anchor=north west, minimum width=1.95cm,
      text width=1.8cm] (h1) at (8,-3.7)
  {\textbf{Constraints}\\
   GNN only \\ 3 CPU hrs\\
   seed=25 \\ Without LLMs \\ ${\leq}$5 runs};
\node[infobox, anchor=north west, minimum width=2.1cm,
      text width=1.95cm] (h2) at (11,-3.7)
  {\textbf{Participants}\\
   13–17 per task\\
   1 final submission};
\node[infobox, anchor=north west, minimum width=2.1cm,
      text width=2.2cm] (h3) at (11,-4.7)
  {\textbf{Output}\\
   \texttt{predictions.csv}\\
   (encrypted)};

\node[hdr={GNNblue}, anchor=west] (p3hdr) at (7.2,-6)
  {Protocol 3 \,·\, LLM Evaluation};

\node[infobox, anchor=north west, minimum width=1.4cm,
      text width=1cm] (l1) at (7.2,-6.4)
  {\textbf{Slot fill}\\
   README\\
   tree · data\\
   spec · reqs};
\node[infobox, anchor=north west, minimum width=1.55cm,
      text width=1cm] (l2) at (8.75,-6.4)
  {\textbf{Frozen}\\
   \textbf{prompt}\\
   zero-shot\\
   no tuning};
\node[infobox, anchor=north west, minimum width=1.55cm,
      text width=1.2cm] (l3) at (10.45,-6.4)
  {\textbf{Plan-then-}\\
   \textbf{code}\\
   $\langle$plan$\rangle$\\
   $\langle$code$\rangle$};
\node[infobox, anchor=north west, minimum width=1.55cm,
      text width=1.2cm] (l4) at (12.15,-6.4)
  {\textbf{Repair}\\
   \textbf{loop}\\
   $k{\leq}5$ iters\\
   traceback};

\node[infobox, anchor=west, minimum width=6.8cm,
      text width=6.cm, font=\tiny, align=center]
  at (7.2,-8)
  {Agent mode (autonomous repo exploration) \quad\textbar\quad
   Non-agent mode (manual prompt filling)};

\node[font=\tiny, text=gray] at (10.6,-8.5)
  {\textit{Evaluated models}};

\foreach \m [count=\i from 0] in {Claude Opus, Gemini 2.5, GPT 5.4, Gemini 3 flash}{
  \node[infobox, anchor=west, font=\tiny]
    at ({7.2 + \i * 1.72}, -8.8) {\m};
}

\foreach \m [count=\i from 0] in {DeepSeek v4, Llama 3.3, Qwen 2.5, Kimi k 2.6}{
  \node[infobox, anchor=west, font=\tiny]
    at ({7.2 + \i * 1.72}, -9.2) {\m};
}


\draw[arr, dashed, gray!50]
  (3.5,-9.6) -- (3.5,-10.) -- (7,-10.5);
\draw[arr, dashed, gray!50]
  (10.6,-9.6) -- (10.6,-10.) -- (8,-10.5);

\node[terminal={GNNpurple}, minimum width=8cm] (eval) at (7,-10.95)
  {Unified hidden-test evaluation pipeline\\
   {\scriptsize GitHub Actions $\cdot$ automated scoring
    $\cdot$ identical conditions for all submissions}};

\draw[arr] (eval.south) -- ++(0,-0.6);

\node[terminal={GNNteal}, minimum width=7cm] (leader)
  at (7,-12.3)
  {Leaderboard + per-task metrics\\
   {\scriptsize human vs.\ LLM $\cdot$ live updates}};


\end{tikzpicture}%
}
\caption{Overview of \textsc{GNN-CB}.
}
\label{fig:overview}
\end{figure*}

Large language models (LLMs) have recently demonstrated strong capabilities in coding and reasoning tasks, including program synthesis, competitive programming, software engineering, and autonomous code repair \citep{chen2021evaluating,li2022codecontests,jimenez2023swe,zhuo2024bigcodebench}. However, existing evaluations mainly focus on general-purpose programming while largely overlooking graph-structured machine learning workflows.

Graph Neural Networks (GNNs) are widely used for learning over graph-structured data in domains such as molecular modeling, recommender systems, social networks, neuroimaging, and knowledge graphs \citep{gilmer2017neural,hamilton2017inductive,li2018diffusion,ying2018graph,li2021, dauparas, Schlichtkrull}. Unlike conventional coding tasks, GNN problems require understanding graph topology, building relational pipelines, selecting message-passing architectures, and optimizing under task-specific constraints, making them more challenging than isolated algorithmic tasks.

Despite the growing importance of graph machine learning, no existing benchmark evaluates whether LLMs can autonomously solve realistic end-to-end GNN coding tasks. Benchmarks such as HumanEval \citep{chen2021evaluating}, CodeContests \citep{li2022codecontests}, SWE-bench \citep{jimenez2023swe}, and BigCodeBench \citep{zhuo2024bigcodebench} focus on algorithmic reasoning or software engineering workflows without graph-learning components, while graph benchmarks such as OGB \citep{hu2020open} and LRGB \citep{dwivedi2022long} are designed to evaluate the predictive performance of GNN models rather than their implementation.

To address this gap, we introduce \textbf{\href{https://github.com/basiralab/GNN-CB}{GNN-CB}}, the first competition-based benchmark designed to evaluate both humans and LLMs on GNN coding tasks.  As illustrated in Figure~\ref{fig:overview}, GNN-CB contains 18 \emph{human-curated} competitions spanning node-, edge-, and graph-level prediction tasks across multiple graph categories, domains, and difficulty tiers. Each competition provides standardized repositories, datasets, hidden test sets, and automated GitHub-based evaluation pipelines.

To ensure fair and reproducible assessment, we define three complementary protocols: (i) competition design, (ii) human evaluation, and (iii) LLM evaluation. Human participants solve tasks under controlled competition constraints with fixed computational budgets and limited submissions. LLMs are evaluated using a frozen zero-shot prompting protocol based on a structured plan-then-code paradigm with bounded execute-and-repair loops. The benchmark additionally supports both non-agent and autonomous agent-based evaluation settings within the same protocol.

Our contributions are summarized as follows. (i) \textbf{\href{https://basiralab.github.io/GNN-CB/}{GNN-CB}} is introduced as the first competition-based benchmark for evaluating both humans and LLMs on GNN coding tasks. (ii) The benchmark contains 18 human-curated competitions spanning diverse graph categories, application domains, and difficulty tiers under unified hidden-test evaluation protocols. (iii) A controlled LLM evaluation framework is proposed based on frozen prompting, plan-then-code generation, and bounded execute-and-repair loops, supporting both agent-based and non-agent settings. (iv) Evaluation of human participants and LLMs enables systematic analysis of performance variability, task difficulty, and implementation challenges in realistic GNN coding environments. (v) GNN-CB is released as a living benchmark with automated evaluation infrastructure, reproducible pipelines, and dynamic leaderboards to support future research on the capabilities and limitations of LLMs as autonomous GNN engineers, as well as serving as a structured resource for hands-on GNN practice across progressively diverse graph-learning tasks.

\section{Related Work}
\label{sec:related}

Recent benchmarks have significantly advanced evaluation of LLM-based code generation and reasoning. HumanEval \citep{chen2021evaluating} and CodeContests \citep{li2022codecontests} introduced standardized evaluation for functional correctness and competitive programming. Later works extended evaluation across programming languages and stricter testing settings, including HumanEval-X \citep{peng2024humaneval}, MultiPL-E \citep{cassano2023multipl}, and EvalPlus \citep{liu2023your}. More realistic software engineering benchmarks such as DS-1000 \citep{lai2023ds}, SWE-bench \citep{jimenez2023swe}, BigCodeBench \citep{zhuo2024bigcodebench}, and LiveCodeBench \citep{jain2024livecodebench} evaluate repository-level reasoning, data science workflows, and instruction-following capabilities.

Graph machine learning benchmarks have standardized evaluation for graph neural networks (GNNs). The Open Graph Benchmark (OGB) \citep{hu2020open} provides datasets and evaluation protocols for node-, edge-, and graph-level prediction tasks, while Benchmarking GNNs \citep{dwivedi2023benchmarking} and LRGB \citep{dwivedi2022long} focus on structural generalization and long-range graph dependencies. However, these benchmarks do not assess whether LLMs can perform at a human level on GNN competitions.

\textsc{GNN-CB} bridges these two research directions by introducing a competition-based benchmark for GNN coding tasks, supporting unified evaluation of both human participants and LLMs under controlled and reproducible conditions.

\section{Benchmark Construction}
\label{sec:benchmark}

\textsc{GNN-CB} is a competition-based benchmark for evaluating GNN coding tasks under standardized settings. Unlike traditional code-generation benchmarks centered on isolated functions or short algorithmic problems, \textsc{GNN-CB} targets repository-level graph learning workflows where participants or LLMs must build complete pipelines spanning data processing, graph construction, architecture design, optimization, and hidden-test prediction generation.

The benchmark construction consists of three protocols : (i) competition design protocol, (ii) human evaluation protocol, and (iii) LLM evaluation protocol.

\subsection{Competition Design Protocol}
\label{subsec:competition_design}

\begin{table*}[t]
\centering
\caption{Overview of all 18 GNN Coding Benchmark competitions.
\textbf{Cat.}~= Graph Category;
\textbf{Diff.}~= Difficulty;
\textbf{N} = typical number of nodes;
\textbf{E} = typical number of edges;
\textbf{Feat.}~= node feature dimensionality. For classification tasks, we report Macro-F1, Accuracy, and Area Under the Curve (AUC), depending on the competition specification. For regression tasks, we use Mean Absolute Error (MAE) for single-target regression and Normalized Mean Absolute Error (NMAE) for multi-target regression.}
\label{tab:overview}

\resizebox{\textwidth}{!}{
\setlength{\tabcolsep}{3pt}
\renewcommand{\arraystretch}{1}
\Large
\begin{tabular}{ccp{6.5cm}cccccccl}
\toprule
\textbf{\#} & Competition & Task Description & Cat. & Diff. & N & E & Feat. & Domain & Metric &
Dataset \\
\midrule

C01 & PROVEN-GNN &
  Graph classification on source code representations &
  \catbox{catHetero}{Hetero} & \diffbox{GNNgold}{Med} &
  260K & 1.7M & 527 & Cybersecurity & Macro-F1 & DiverseVul \\

 C02 & CGCC &
   Graph classification of city street-network layouts &
   \catbox{catHomo}{Homo} & \diffbox{GNNred}{Hard} &
   9.396K & 13.533K & 3 & Urban &
   Macro-F1 & OpenStreetMap \\

C03 & Real Or Fake! &
  Binary graph classification for fake news detection &
  \catbox{catHier}{Hier} & \diffbox{GNNgreen}{Easy} &
  314K & 308K & 1078 & Social & Binary F1 & GossipCop \\

C04 & GLIMPS-GNN &
  Inductive cfRNA $\to$ preeclampsia classification &
  \catbox{catGen}{Gen} & \diffbox{GNNred}{Hard} &
  320 & 3200 & 6650 & Biomedical & F1 & GEO \\

C05 & GNN\_BACE &
  Molecular graph classification for BACE-1 inhibition &
  \catbox{catMol}{Mol} & \diffbox{GNNgold}{Med} &
  32 & 34 & 8 & Medical & Macro-F1 & BACE (MoleculeNet) \\

C06 & DiaGraph &
  Binary node classification for diabetes prediction &
  \catbox{catHomo}{Homo} & \diffbox{GNNred}{Hard} &
  96K & 1.2M & 12 & Medical & Macro-F1 & Kaggle Diabetes \\

C07 & GRAPE &
  Binary graph classification for diabetic retinopathy &
  \catbox{catSpatial}{Spatial} & \diffbox{GNNgold}{Med} &
  252 & 846 & 4 & Medical & Macro-F1 & DRIVE/STARE/HRF \\

C08 & Liar Nodes &
  Node Classification under Adversarial Feature Corruption for Cancer Single-Cell Analysis &
  \catbox{catHomo}{Homo} & \diffbox{GNNgreen}{Easy} &
  3.7K & 27K & 1.7K & Biomedical & Accuracy & CancerSCEM \\

C09 & GraphFMD &
  Temporal inductive node classification (Bitcoin) &
  \catbox{catTemporal}{Temp} & \diffbox{GNNgold}{Med} &
  25.5K & --- & 165 & Finance & Macro-F1 & Elliptic \\

C10 & NetLinkArena &
  Link prediction in sparse citation graphs &
  \catbox{catHomo}{Homo} & \diffbox{GNNgold}{Med} &
  3.3K & 9.1K & 2.7K & Academic & AUC & Citation \\

C11 & PARK-GNN &
  Node classification for Parkinson's detection &
  \catbox{catHomo}{Homo} & \diffbox{GNNred}{Hard} &
  195 & $\sim$1K & 22 & Medical & Macro-F1 & UCI Parkinson's \\

C12 & GNN-CoRA &
  Node classification on citation network &
  \catbox{catHomo}{Homo} & \diffbox{GNNgold}{Med} &
  2.7K & 5.4K & 1433 & Academic & Accuracy & Cora \\

C13&{GTA}&
  Graph classification on ideal and perturbed conditions &
  \catbox{catGen}{Gen} & \diffbox{GNNred}{Hard} &
  $\sim$18 & $\sim$ 40 &7& Bioinformatics&Macro-F1& {{MUTAG}} \\

C14 & BrainAge &
  Graph regression for age prediction from brain connectome &
  \catbox{catHomo}{Homo} & \diffbox{GNNgreen}{Easy} &
  29.3K & 478.8K & 2 & Neuroimaging & MAE & OASIS-3 \\

C15 & OctoNode Cup &
  Node classification on GitHub social network &
  \catbox{catHomo}{Homo} & \diffbox{GNNgold}{Med} &
  37.7K & 289K & 42 & Social & Macro-F1 & GitHub Social \\

C16 & Graph4ASD &
  Binary classification of brain connectivity graphs &
  \catbox{catHomo}{Homo} & \diffbox{GNNgold}{Med} &
  127.4K & 12676.3K & 1 & Neuroimaging & Macro-F1 & ABIDE \\

C17 & Mix2Print &
  Learning material interactions for 3D bioprinting &
  \catbox{catHomo}{Homo} & \diffbox{GNNred}{Hard} &
  423 & 740 & 442 & Bioprinting & Normalized-MAE & CECT \\

C18 & THInC &
  Inductive GNN for cell type classification &
  \catbox{catSpatial}{Spatial} & \diffbox{GNNred}{Hard} &
  51K & 258K & 4 & Biomedical & Macro-F1 & NuCLS \\

\bottomrule
\end{tabular}%
}
\end{table*}

The competition design protocol organizes benchmark tasks across diverse graph learning settings. The benchmark contains 18 independent competitions spanning node classification, edge prediction, graph classification, and graph regression tasks (Figure~\ref{fig:overview}). Competitions are organized along two axes: graph category and difficulty tier. The benchmark covers seven graph categories, including \textcolor{catHomo}{homogeneous}, \textcolor{catHetero}{heterogeneous}, \textcolor{catTemporal}{temporal}, \textcolor{catSpatial}{spatial}, \textcolor{catHier}{hierarchical}, \textcolor{catMol}{molecular}, and \textcolor{catGen}{generalization} graphs. Competitions are additionally grouped into \colorbox{GNNgreen!30}{\strut\,Easy}, \colorbox{GNNgold!30}{\strut\,Medium} and \colorbox{GNNred!30}{\strut\,Hard} difficulty tiers based on implementation complexity, architectural requirements, and observed human performance. Table~\ref{tab:overview} summarizes all benchmark competitions, including graph categories, dataset scales, feature dimensionalities, application domains, and evaluation metrics. Detailed dataset specifications and repository descriptions are provided in Appendix~\ref{app:specs}.

Each competition is associated with a dedicated repository that specifies the task, submission format, evaluation metrics, and leaderboard configuration. An automated evaluation infrastructure is provided for submission validation, scoring, leaderboard updates, and execution logging, ensuring that all submissions are processed through a unified evaluation workflow across competitions. 

All submissions are evaluated exclusively on hidden test sets that remain inaccessible throughout the competition lifecycle. Evaluation is performed through automated GitHub-based pipelines that validate submission formats, execute scoring scripts, and update leaderboards without manual intervention.

The infrastructure additionally supports secure submission mechanisms to prevent information leakage. Depending on repository configuration, participants either submit predictions through controlled external platforms or upload encrypted prediction files to designated repository directories, where they are automatically decrypted only during evaluation prior to scoring.


\subsection{Human Evaluation Protocol}
\label{subsec:human_protocol}

The human evaluation protocol measures participant performance
under standardized computational and submission constraints.
Participants solve each competition independently using only
GNN models, without external solutions, and submit a single
final prediction file evaluated through the same automated
hidden-test pipeline used for LLMs. LLM usage during the human track is restricted to low-level utility support without exposure to task-specific benchmark information. 
More detailed protocol and participant information are provided in Appendix~\ref{app:human_eval}.
Participants represent 18 universities across 12 countries (Figure~\ref{fig:participants_world}, Appendix~\ref{app:human_eval}), ensuring geographic diversity in benchmark design and human performance evaluation.

\subsection{LLM Evaluation Protocol}
\label{sec:llm_protocol}

GNN-CB evaluates LLMs under a controlled and reproducible protocol designed to measure repository-grounded graph learning ability rather than prompt-engineering performance. All models are evaluated using a frozen prompt template shared across all competitions, without task-specific tuning or adaptive prompting \citep{zhuo2024bigcodebench}. As illustrated in Figure~\ref{fig:compact_pipeline}, repository information is converted into a standardized prompt through slot filling \citep{jain2024livecodebench}, after which the model generates a structured reasoning plan and an executable code following a plan-then-code paradigm \citep{jiang2024self}. The generated solution is executed locally in a controlled environment and optionally refined through bounded execute-and-repair iterations \citep{yang2024swe}. Final predictions are evaluated on hidden test sets using the same automated infrastructure used for human participants, enabling direct and fair comparison under identical conditions. The protocol consists of four components: (i) frozen prompt instantiation, (ii) plan-then-code generation, (iii) execution and repair, and (iv) evaluation settings.


\tikzset{
  pbox/.style 2 args={
    draw=#1!60, fill=#1!10, text=#1!80!black,
    rounded corners=5pt,
    minimum width=#2, minimum height=0.6cm,
    align=center, inner sep=5pt, font=\scriptsize
  },
  ibox/.style={pbox={GNNblue}{#1}},     ibox/.default=1.9cm,
  gbox/.style={pbox={GNNpurple}{#1}},   gbox/.default=1.9cm,
  tbox/.style={pbox={GNNteal}{#1}},     tbox/.default=1.9cm,
  ebox/.style={pbox={GNNgreen}{#1}},    ebox/.default=1.9cm,
  abox/.style={pbox={GNNorange}{#1}},   abox/.default=1.9cm,
  rbox/.style={pbox={GNNgold}{#1}},     rbox/.default=1.9cm,
  dnode/.style={
    draw=GNNpurple!60, fill=GNNpurple!10,
    text=GNNpurple!80!black,
    shape=diamond, aspect=2.2,
    minimum width=1.8cm, minimum height=0.6cm,
    inner sep=2pt, font=\scriptsize
  },
  ppill/.style={
    draw=#1!50, fill=#1!10, text=#1!70!black,
    rounded corners=3pt,
    font=\scriptsize\bfseries,
    inner xsep=6pt, inner ysep=2pt
  },
  stag/.style={
    draw=GNNpurple!35, fill=GNNpurple!8,
    text=GNNpurple!80!black,
    rounded corners=3pt,
    font=\tiny\ttfamily,
    minimum width=2.5cm, minimum height=0.38cm,
    inner xsep=4pt, inner ysep=1.5pt
  },
  blarr/.style={-{Stealth[length=4pt]}, thick, draw=GNNblue!65},
  purrarr/.style={-{Stealth[length=4pt]}, thick, draw=GNNpurple!65},
  tealarr/.style={-{Stealth[length=4pt]}, thick, draw=GNNteal!65},
  orarr/.style={-{Stealth[length=4pt]}, thick, draw=GNNorange!75},
  greyarr/.style={-{Stealth[length=4pt]}, thick, draw=GNNgray!55},
}

\tcbset{
  plansty/.style={
    enhanced, sharp corners=downhill,
    arc=6pt, boxrule=0.55pt,
    colframe=GNNpurple!55, colback=GNNpurple!5,
    fonttitle=\scriptsize\bfseries, coltitle=GNNpurple!75!black,
    attach boxed title to top left={yshift=-2.2mm, xshift=5mm},
    boxed title style={
      sharp corners, arc=3pt, boxrule=0.4pt,
      colback=white, colframe=GNNpurple!50,
    },
    left=5pt, right=5pt, top=6pt, bottom=4pt,
  },
  codesty/.style={
    enhanced, sharp corners=downhill,
    arc=6pt, boxrule=0.55pt,
    colframe=GNNteal!55, colback=GNNteal!4,
    fonttitle=\scriptsize\bfseries, coltitle=GNNteal!75!black,
    attach boxed title to top left={yshift=-2.2mm, xshift=5mm},
    boxed title style={
      sharp corners, arc=3pt, boxrule=0.4pt,
      colback=white, colframe=GNNteal!50,
    },
    left=5pt, right=5pt, top=6pt, bottom=4pt,
  },
}

\lstdefinestyle{GNNcode}{
  language=Python,
  basicstyle=\tiny\ttfamily,
  keywordstyle=\color{GNNblue}\bfseries,
  commentstyle=\color{GNNgray!65}\itshape,
  stringstyle=\color{GNNorange!80},
  emphstyle=\color{GNNteal}\bfseries,
  emph={SAGEConv,Adam,BCEWithLogitsLoss,StandardScaler,
        global_mean_pool,global_max_pool},
  numbers=none,
  breaklines=true,
  showstringspaces=false,
  frame=none,
  aboveskip=0pt, belowskip=0pt,
  xleftmargin=0pt,
}


\begin{figure*}[t]
\centering
\scalebox{0.85}{%
\begin{tikzpicture}[
  node distance=0.4cm and 0.35cm,
  every node/.style={font=\tiny},
]

\node[ppill=GNNblue]   (ph1) at (0,0)     {Input};
\node[ppill=GNNpurple] (ph2) at (4.3,0)  {Generation};
\node[ppill=GNNteal]   (ph3) at (7.8,0)  {Execution};
\node[ppill=GNNorange] (ph4) at (13.8,0) {Scoring};

\foreach \x in {3.3, 6.9, 12.7}{
  \draw[GNNgray!20, dashed, thin] (\x,0.22) -- (\x,-3.6);
}

\node[ibox=2.1cm, below=0.2cm of ph1] (repo)
  {\textbf{Repository}\\[1pt]
   {\tiny README}\\{\tiny Data}\\{\tiny Tree}\\{\tiny Submission Spec.}};

\node[ibox=1.75cm, right=0.32cm of repo] (slots)
  {\textbf{Slot filling}\\[1pt]
   {\tiny 5 slots}};

\node[gbox=1.85cm, right=0.32cm of slots] (prompt)
  {\textbf{Frozen}\\[-2pt]\textbf{prompt}\\[1pt]
   {\tiny zero-shot}};

\node[gbox=1.1cm, right=0.32cm of prompt] (llm)
  {\textbf{LLM}};

\node[tbox=1.6cm, right=0.32cm of llm,
      minimum height=0.42cm, inner sep=2pt,
      font=\tiny\bfseries] (plan)
  {$\langle\texttt{plan}\rangle$ \\ $\langle\texttt{code}\rangle$};

\node[ebox=1.6cm, right=0.32cm of plan] (exec)
  {\textbf{Execute}\\[1pt]
   {\tiny CPU-only}};

\node[dnode, right=0.32cm of exec] (err) {error?};

\node[abox=1.75cm, right=0.32cm of err] (submit)
  {\textbf{Submit}\\[1pt]
   {\tiny \texttt{predictions.csv}}};

\node[abox=2.9cm, below=0.3cm of submit] (score)
  {\textbf{Hidden-test scoring}\\[1pt]
   {\tiny task-specific metric}};

\node[abox=2.7cm, below=0.3cm of score] (leader)
  {\textbf{Leaderboard}\\[1pt]
   {\tiny live update}};

\node[rbox=4.0cm, left=1.5cm of score] (repair)
  {\textbf{Repair loop}\enspace($k\leq 5$)\\[1pt]
   {\tiny traceback $\to$ LLM $\to$ re-generate}};

\draw[blarr]   (repo)   -- (slots);
\draw[blarr]   (slots)  -- (prompt);
\draw[purrarr] (prompt) -- (llm);
\draw[purrarr] (llm)    -- (plan);
\draw[tealarr] (plan)   -- (exec);


\draw[tealarr] (exec) -- (err);

\draw[purrarr] (err) --
  node[above, font=\scriptsize, text=GNNgray!70] {no}
  (submit);

\draw[purrarr] (err.south) -- ++(0,-0.98)
  node[right=1pt, font=\scriptsize, text=GNNgray!70,
       yshift=0.12cm] {yes}
  -| (repair.east);

\draw[purrarr] (repair.west) -- ++(-1.05,0) -- (llm.south);

\draw[orarr] (submit.south) -- (score.north);

\draw[orarr] (score) -- (leader);

\node[font=\scriptsize, text=GNNgray!65,
      above=0.5cm] at ($(repo.east)!0.5!(slots.west)$) {};
\node[font=\scriptsize, text=GNNgray!65,
      above=0.5cm] at ($(llm.east)!0.5!(plan.west)$)   {};

\end{tikzpicture}
}%
\caption{LLM evaluation pipeline.}
\label{fig:compact_pipeline}
\end{figure*}
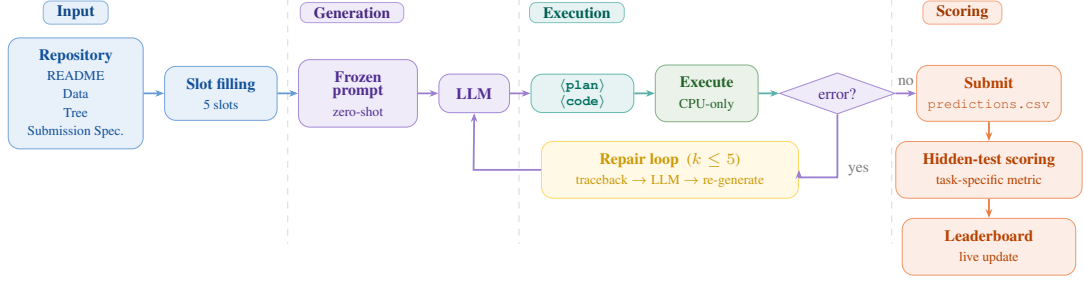

\subsubsection{Frozen Prompt Instantiation}
\label{subsec:frozen_prompt}


All models were evaluated using a standardized prompt template with limited, controlled task-specific context added to ensure executable outputs. The template contains predefined slots populated with repository-derived information such as the README, repository tree, dataset summary, submission specification, and dependency requirements.

Repository components are inserted into structured slots without manual modification \citep{jain2024livecodebench}. The instantiated prompt requires the model to produce a structured \texttt{<plan>} followed by executable \texttt{<code>}, providing a standardized interface across all competitions. Figure~\ref{fig:prompt_instantiation} in Appendix~\ref{app:llm_eval} illustrates the slot-filling construction, and the full template is provided in Listing~\ref{lst:frozen_prompt}.

\subsubsection{Plan-Then-Code Generation}
\label{subsec:plan_then_code}

Given the instantiated prompt, the model generates: (i) a high-level reasoning plan and (ii) an executable code. This plan-then-code strategy \citep{jiang2024self} encourages explicit reasoning before implementation and reduces common failures such as invalid pipelines, hallucinated APIs, or malformed submissions. The generated plan typically specifies feature engineering, GNN architecture design, optimization procedures, validation strategy, and submission generation. The accompanying code implements the complete pipeline as a standalone executable solution. A representative plan-then-code output is shown in Figure~\ref{fig:plan_block} in Appendix~\ref{app:llm_eval}.









\subsubsection{Execution and Repair Loop}
\label{subsec:repair_loop}

Generated code is executed locally in a controlled environment under CPU-only constraints. If execution fails due to syntax errors, runtime exceptions, dependency issues, or invalid outputs, an execute-and-repair loop \citep{yang2024swe} is triggered in which the execution traceback is appended to the conversation context and returned to the model for refinement. Repair attempts are bounded ($k \leq 5$) to maintain a controlled evaluation budget. 

\subsubsection{Agent-Based and Non-Agent Evaluation}
\label{subsec:agent_vs_nonagent}

\textsc{GNN-CB} supports two complementary execution modes: \emph{non-agent} and \emph{agent-based}. In the \emph{non-agent} mode, a lightweight orchestration script instantiates the frozen prompt with repository-derived slots, dispatches the API request, parses the response into \texttt{<plan>} and \texttt{<code>} blocks, executes the resulting \texttt{solution.py} under the CPU-only environment, and feeds runtime tracebacks back into the prompt for a bounded number of repair iterations; a simplified version of this control loop is reproduced in Listing~\ref{lst:nonagent_loop} of Appendix~\ref{app:llm_eval}. In the \emph{agent-based} mode, an autonomous coding system inspects repository contents, generates an executable solution, runs it, and iteratively repairs failures without human intervention, using the same execute-and-repair budget. Both modes consume identical repository information, enforce the same execution constraints, and submit through the same automated hidden-test pipeline, so that observed performance differences reflect the model's reasoning and implementation ability rather than prompt-engineering effort or orchestration design.

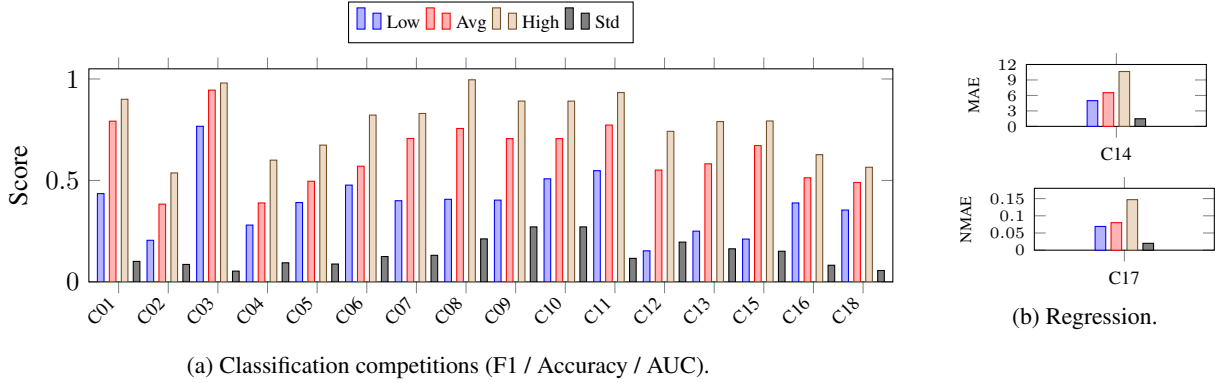
\begin{figure*}[h!]
\centering

\begin{minipage}{0.74\textwidth}
\centering
\begin{tikzpicture}
\begin{axis}[
    width=\textwidth,
    height=4.4cm,
    ybar,
    bar width=2.5pt,
    ymin=0, ymax=1.05,
    xtick=data,
    symbolic x coords={C01,C02,C03,C04,C05,C06,C07,C08,C09,C10,C11,C12,C13,C15,C16,C18},
    xticklabels={C01,C02,C03,C04,C05,C06,C07,C08,C09,C10,C11,C12,C13,C15,C16,C18},
    xticklabel style={rotate=45, anchor=east, font=\scriptsize},
    ylabel={Score},
    legend style={at={(0.5,1.12)},anchor=south,legend columns=4, font=\scriptsize},
    enlarge x limits=0.04,
]
\addplot coordinates {
(C01,0.435) (C02,0.205) (C03,0.767) (C04,0.28)
(C05,0.391) (C06,0.477) (C07,0.40)  (C08,0.407)
(C09,0.403) (C10,0.508) (C11,0.548) (C12,0.153)
(C13,0.25)  (C15,0.211) (C16,0.389) (C18,0.354)
};
\addplot coordinates {
(C01,0.792) (C02,0.383) (C03,0.945) (C04,0.389)
(C05,0.496) (C06,0.570) (C07,0.707) (C08,0.756)
(C09,0.706) (C10,0.706) (C11,0.773) (C12,0.551)
(C13,0.582) (C15,0.672) (C16,0.513) (C18,0.490)
};
\addplot coordinates {
(C01,0.900) (C02,0.537) (C03,0.980) (C04,0.600)
(C05,0.674) (C06,0.822) (C07,0.830) (C08,0.996)
(C09,0.891) (C10,0.891) (C11,0.933) (C12,0.742)
(C13,0.790) (C15,0.793) (C16,0.627) (C18,0.565)
};
\addplot coordinates {
(C01,0.101) (C02,0.086) (C03,0.053) (C04,0.094)
(C05,0.088) (C06,0.125) (C07,0.131) (C08,0.212)
(C09,0.271) (C10,0.271) (C11,0.116) (C12,0.196)
(C13,0.163) (C15,0.151) (C16,0.082) (C18,0.056)
};
\legend{Low, Avg, High, Std}
\end{axis}
\end{tikzpicture}

\vspace{2pt}
{\small (a) Classification competitions (F1 / Accuracy / AUC).}
\end{minipage}%
\hfill
\begin{minipage}{0.24\textwidth}
\centering

\begin{tikzpicture}
\begin{axis}[
    width=\textwidth,
    height=2.4cm,
    ybar,
    bar width=4pt,
    ymin=0, ymax=12,
    ytick={0, 3, 6, 9, 12},
    scaled y ticks=false,
    xtick=data,
    symbolic x coords={C14},
    xticklabel style={font=\scriptsize},
    ylabel={MAE},
    ylabel style={font=\tiny},
    yticklabel style={/pgf/number format/fixed, /pgf/number format/precision=0, font=\tiny},
    title style={font=\scriptsize},
    enlarge x limits=0.9,
]
\addplot coordinates {(C14,4.984)};   
\addplot coordinates {(C14,6.531)};   
\addplot coordinates {(C14,10.638)};  
\addplot coordinates {(C14,1.474)};   
\end{axis}
\end{tikzpicture}

\vspace{2pt}

\begin{tikzpicture}
\begin{axis}[
    width=\textwidth,
    height=2.4cm,
    ybar,
    bar width=4pt,
    ymin=0, ymax=0.18,
    ytick={0, 0.05, 0.10, 0.15},
    scaled y ticks=false,
    xtick=data,
    symbolic x coords={C17},
    xticklabel style={font=\scriptsize},
    ylabel={NMAE},
    ylabel style={font=\tiny},
    yticklabel style={/pgf/number format/fixed, /pgf/number format/precision=2, font=\tiny},
    title style={font=\scriptsize},
    enlarge x limits=0.9,
]
\addplot coordinates {(C17,0.069)};   
\addplot coordinates {(C17,0.08)};    
\addplot coordinates {(C17,0.147)};   
\addplot coordinates {(C17,0.02)};    
\end{axis}
\end{tikzpicture}

\vspace{1pt}
{\small (b) Regression.}
\end{minipage}

\caption{Per-task human-performance statistics across all 18
\textsc{GNN-CB} competitions: minimum, average, maximum, and
standard deviation of participant scores. Panel~(a) covers the
16 classification competitions on a common 0--1 axis (F1 /
Accuracy / AUC); panel~(b) covers the two regression
competitions (C14 BrainAge, MAE; C17 Mix2Print, NMAE) on
independent axes due to different metric scales. Higher is
better in (a); lower is better in (b).}
\label{fig:human_results}

\end{figure*}

\subsubsection{LLM Implementations}
\label{sec:implementation_details}

To ensure deterministic, reproducible outputs, all models are decoded with \texttt{temperature = 0}, applied where the provider's API exposes this setting. API-only (non-agent) runs use a per-call output budget of \texttt{max\_output\_tokens = 10K}, while agent-mode runs follow each agent's native session budget, which may consume substantially more tokens across multiple internal tool calls and extended-thinking traces. A full inventory of evaluated models, their API identifiers, and source links is provided in Table~\ref{tab:llms} (Appendix~\ref{app:llm_eval}).

In the API-only mode, models are accessed through provider SDKs or OpenAI-compatible endpoints, including Together AI serverless inference, and receive the frozen prompt through a lightweight orchestration script that handles slot filling, code extraction, execution, and bounded repair iterations. Models evaluated in this mode include Gemini-3 Flash, Gemini 2.5 Pro, Llama-3.3 70B, Claude Opus 4.6, Qwen2.5-Coder 32B, Kimi k2.6, and DeepSeek v4 Pro.

In the agent-based mode, models operate as autonomous coding agents that directly inspect repository contents, generate solutions, execute code, and iteratively repair failures without human intervention. Models evaluated in this mode include Claude Opus 4.7 and GPT-5.4 Codex.

All generated solutions are executed locally under CPU-only constraints inside isolated environments. The framework automatically parses model outputs into \texttt{<plan>} and \texttt{<code>} sections, executes the generated \texttt{solution.py}, performs bounded repair iterations ($k\!\leq\!5$), validates prediction files, and logs execution artifacts including runtime traces and generated submissions. This design enables reproducible evaluation across multiple model families without requiring local GPU deployment.

\section{Results and Analysis}
\label{sec:human}

We evaluate \textsc{GNN-CB} along two complementary tracks: a human-participant track (Section~\ref{subsec:human_protocol}) with expert volunteers under standardized reproducibility constraints, and an LLM track (Section~\ref{sec:llm_protocol}) with nine proprietary and open-source models under a frozen plan-then-code protocol with bounded execute-and-repair iterations. Both tracks use the same automated hidden-test pipeline and task-specific metrics: Macro-F1, accuracy, and AUC for classification, and MAE/NMAE for regression. We report human-cohort statistics, each LLM's per-competition score and gap to the human top, and cross-task variability as a measure of consistency.

\begin{table}[h!]
\centering
\scriptsize
\setlength{\tabcolsep}{2.5pt}
\renewcommand{\arraystretch}{1.05}

\caption{Descriptive statistics of human performance across competitions. Difficulty level (Diff), number of participants (P), average score (AVG), and standard deviation (STD) summarize overall human performance.High and Low columns report the maximum and minimum human scores observed, respectively, Metric indicates the evaluation criterion used for tasks.}
\label{tab:human}

\begin{tabular}{clccccccl}
\toprule

\# & Competition & Diff. & P & AVG & STD & High & Low & Metric  \\
\midrule
C01 & PROVEN-GNN & Med & 17 & 0.792 & 0.101 & 0.900 & 0.435 & F1 \\
C02 & CGCC & Hard & 17 & 0.383 & 0.086 & 0.537 & 0.205 & F1\\
C03 & Real Or Fake! & Easy & 15 & 0.945 & 0.053 & 0.980 & 0.767 & F1\\
C04 & GLIMPS-GNN & Hard & 16 & 0.389 & 0.094 & 0.600 & 0.28 & F1\\
C05 & GNN\_BACE & Med & 14 & 0.496 & 0.088 & 0.674 & 0.391 & F1\\
C06 & DiaGraph & Hard & 15 & 0.570 & 0.125 & 0.822 & 0.477 & F1\\
C07 & GRAPE & Med & 16 & 0.707 & 0.131 & 0.830 & 0.4 & F1\\
C08 & Liar Nodes & Easy &17  & 0.756 & 0.212  & 0.996 & 0.407& Acc.\\
C09 & GraphFMD & Med & 17 & 0.706 & 0.271 & 0.891 & 0.403 & F1\\
C10 & NetLinkArena & Med & 17 & 0.706 & 0.271 & 0.891 & 0.508 & AUC\\
C11 & PARK-GNN &  Hard & 14 & 0.773 & 0.116 & 0.933 & 0.548 & F1\\
C12 & GNN-CoRA & Med & 14 & 0.551 & 0.196 & 0.742 & 0.153& Acc.\\
C13&GTA&Hard&17& 0.5816 & 0.1629 & 0.79 & 0.25 & F1\\
C14 & BrainAge & Easy & 14 & 6.531 & 1.474 & 10.638 &  4.984 & MAE\\
C15 & OctoNode Cup & Med & 15 & 0.672 & 0.151 & 0.793 & 0.211 & F1\\
C16 & Graph4ASD & Med & 14 & 0.513 & 0.082 & 0.627 & 	0.389 &  F1\\
C17 & Mix2Print & Hard & 13 & 0.08\ & 0.02 & 0.147 & 0.069 & NMAE\\
C18 & THInC & Hard & 14 & 0.490 & 0.056 & 0.565 & 0.354 & F1\\
\midrule
\end{tabular}
\end{table}

\definecolor{redcell}{HTML}{F5B7B1}
\definecolor{redcellSm}{HTML}{FADBD8}
\definecolor{greencell}{HTML}{CDEBC4}
\definecolor{yellowcell}{HTML}{FFF3B0}
\definecolor{graycell}{HTML}{FAFAFA}
\definecolor{humanrow}{HTML}{F7F5E8}
\definecolor{gaprow}{HTML}{EEEAF5}
\definecolor{groupband}{HTML}{FBFAF3}
\definecolor{gapblue}{HTML}{1F4E8B}
\definecolor{gapgreen}{HTML}{1B6B2E}
\definecolor{diffEasyBg}{HTML}{4A8A4F}
\definecolor{diffMedBg}{HTML}{B28A2C}
\definecolor{diffHardBg}{HTML}{B3463D}

\newcommand{\Nb}[1]{#1}                       
\newcommand{\B}[1]{\textbf{#1}}               
\newcommand{\U}[1]{\underline{#1}}            
\newcommand{\BU}[1]{\textbf{\underline{#1}}}  

\newcommand{\cd}[2]{\cellcolor{redcell}{#1}$_{\textcolor{gapblue}{\scriptstyle\downarrow\!#2}}$}
\newcommand{\cdL}[2]{\cellcolor{redcellSm}{#1}$_{\textcolor{gapblue}{\scriptstyle\downarrow\!#2}}$}
\newcommand{\cu}[2]{\cellcolor{greencell}{#1}$_{\textcolor{gapgreen}{\scriptstyle\uparrow\!#2}}$}
\newcommand{\cz}[1]{\cellcolor{yellowcell}{#1}$_{\scriptstyle 0}$}
\newcommand{\cna}{\cellcolor{graycell}\textcolor{gray}{\textemdash}}

\newcommand{\diffBand}[3]{\multicolumn{#1}{c}{\cellcolor{#2}{\color{white}\textbf{\textit{#3}}}}}

\setlength{\tabcolsep}{2.5pt}
\renewcommand{\arraystretch}{1.25}

\pagestyle{empty}\thispagestyle{empty}

\begin{table*}[!t]
\centering
\caption{\textbf{LLM Benchmark Leaderboard on GNN Competitions.}
Scores and gaps to Human Top are reported for 9 LLMs across 18 GNN competitions. Gaps are computed as $\text{score}-\text{human top}$ and sign-corrected for regression.
\textcolor{gapblue}{Blue subscripts $\downarrow$} denote performance below Human Top;
\textcolor{gapgreen}{Green subscripts $\uparrow$} denote ties or improvements.
For classification tasks, cells are tinted \colorbox{redcell}{\strut\,red} ($ -5\%  $ gap $ < 0 $\%),
\colorbox{redcellSm}{\strut\,light red} (gap $<5$\%),
\colorbox{yellowcell}{\strut\,yellow} (tied) and \colorbox{greencell}{\strut\,green} (gap $> 0$). For regression tasks, cells are colored as  \colorbox{redcell}{\strut\,red} and  \colorbox{greencell}{\strut\,green} for below and above Human Top, respectively. 
\underline{Underlining} marks the best result within each model category, and \textbf{boldface} the best overall result per competition.
The final row reports the average human--LLM gap across the nine models.}
\label{tab:gnn-llm-leaderboard-cls}
\small

\begin{adjustbox}{max width=\textwidth}
\begin{tabular}{@{}l@{\hspace{4pt}} *{3}{c} | *{9}{c} | *{6}{c}@{}}
\toprule
\textbf{Models}
  & \diffBand{3}{diffEasyBg}{Easy}
  & \diffBand{9}{diffMedBg}{Medium}
  & \diffBand{6}{diffHardBg}{Hard} \\
\cmidrule(lr){2-4}\cmidrule(lr){5-13}\cmidrule(lr){14-19}
   & \textbf{C03} & \textbf{C08} & \textbf{C14} &\textbf{C01} & \textbf{C05} & \textbf{C07} & \textbf{C09} & \textbf{C10} & \textbf{C11} & \textbf{C12} & \textbf{C15} & \textbf{C16} & \textbf{C02} & \textbf{C04} & \textbf{C06} & \textbf{C13} & \textbf{C17} & \textbf{C18} \\
   & {\scriptsize\itshape Real\_Or\_Fake} & {\scriptsize\itshape Liar Nodes} & 
   {\scriptsize\itshape Brain-Age} & {\scriptsize\itshape PROVEN} & {\scriptsize\itshape GNN\_BACE} & {\scriptsize\itshape GRAPE} & {\scriptsize\itshape Graph FMD} & {\scriptsize\itshape NetLinkArena} & {\scriptsize\itshape PARK-GNN} & {\scriptsize\itshape GNN\_CoRA} & {\scriptsize\itshape OctoNode} & {\scriptsize\itshape Graph4ASD} & {\scriptsize\itshape CGCC} & {\scriptsize\itshape GLIMPS-GNN} & {\scriptsize\itshape DiaGraph} & {\scriptsize\itshape GTA} & 
   {\scriptsize\itshape Mix2Print} & {\scriptsize\itshape THInC} \\
   & {\scriptsize\itshape val F1} & {\scriptsize\itshape acc.} & 
   {\scriptsize\itshape MAE} &{\scriptsize\itshape mF1} & {\scriptsize\itshape F1} & {\scriptsize\itshape mF1} & {\scriptsize\itshape val F1} & {\scriptsize\itshape AUC} & {\scriptsize\itshape val F1} & {\scriptsize\itshape acc.} & {\scriptsize\itshape Mac-F1} & {\scriptsize\itshape Mac-F1} & {\scriptsize\itshape OOF F1} & {\scriptsize\itshape F1} & {\scriptsize\itshape Mac-F1} & {\scriptsize\itshape F1 (P.)} & 
   {\scriptsize\itshape NMAE} & {\scriptsize\itshape Mac-F1} \\
\midrule
\multicolumn{19}{c}{\cellcolor{groupband}\itshape Proprietary Reasoning LLMs (Frontier)} \\[1pt]
\midrule
Gemini 2.5 Pro~\cite{gemini_2_5_pro} & \cdL{\Nb{93.6}}{4.5} & \cd{\Nb{92.7}}{6.9} & \cd{\Nb{5.524}}{0.540} &\cd{\Nb{75.5}}{14.5} & \cd{\BU{61.9}}{5.5} & \cz{\BU{83.0}} & \cu{\U{83.6}}{3.0} & \cdL{\BU{87.5}}{1.6} & \cd{\Nb{71.2}}{22.1} & \cd{\Nb{64.8}}{9.4} & \cd{\Nb{54.1}}{25.2} & \cd{\Nb{48.8}}{13.9} & \cd{\U{40.7}}{13.0} & \cd{\Nb{29.0}}{31.0} & \cd{\Nb{48.9}}{33.4} & \cd{\Nb{65.7}}{14.2} & \cdL{\Nb{0.093}}{0.025} &\cd{\Nb{36.0}}{20.6} \\
Claude Opus 4.7~\cite{claude_opus_4_7} & \cdL{\BU{97.6}}{0.5} & \cu{\BU{99.8}}{0.2} & \cd{\BU{5.504}}{0.520} &\cd{\U{78.1}}{11.9} & \cd{\Nb{55.4}}{12.0} & \cz{\Nb{83.0}} & \cdL{\Nb{76.2}}{4.4} & \cd{\Nb{69.6}}{19.5} & \cdL{\U{90.2}}{3.0} & \cd{\Nb{55.1}}{19.1} & \cdL{\U{74.6}}{4.8} & \cdL{\U{60.8}}{1.9} & \cd{\Nb{32.6}}{21.0} & \cd{\Nb{44.0}}{16.0} & \cd{\U{76.9}}{5.3} & \cdL{\Nb{79.1}}{0.8} & \cdL{\U{0.069}}{0.000} &\cu{\U{57.2}}{0.6} \\
GPT-5.4 Codex~\cite{gpt54_codex} & \cd{\Nb{70.7}}{27.3} & \cd{\Nb{88.9}}{10.6} & \cd{\Nb{5.847}}{0.863} &\cd{\Nb{50.2}}{39.8} & \cd{\Nb{53.4}}{14.0} & \cd{\Nb{44.4}}{38.5} & \cdL{\Nb{77.6}}{3.1} & \cd{\Nb{83.9}}{5.2} & \cd{\Nb{84.6}}{8.7} & \cdL{\U{74.0}}{0.2} & \cd{\Nb{71.8}}{7.5} & \cd{\Nb{45.7}}{17.0} & \cd{\Nb{36.2}}{17.5} & \cd{\BU{52.0}}{8.0} & \cd{\Nb{72.5}}{9.7} & \cu{\BU{83.5}}{3.6} & \cdL{\Nb{0.074}}{0.005} &\cd{\Nb{48.6}}{7.9} \\
\midrule
\multicolumn{19}{c}{\cellcolor{groupband}\itshape Open-source Reasoning LLMs} \\[1pt]
\midrule
DeepSeek-v4 Pro~\cite{deepseek_v4_pro} & \cdL{\U{95.2}}{2.8} & \cd{\U{73.0}}{26.6} & \cd{\U{5.512}}{0.528} & \cd{\U{79.4}}{10.6} & \cd{\U{53.5}}{13.9} & \cz{\U{83.0}} & \cdL{\U{78.1}}{2.5} & \cd{\U{66.5}}{22.7} & \cd{\U{42.6}}{50.6} & \cdL{\U{70.0}}{4.2} & \cd{\U{54.2}}{25.1} & \cd{\U{32.9}}{29.8} & \cd{\U{15.6}}{38.1} & \cd{\U{30.0}}{30.0} & \cd{\U{74.4}}{7.9} & \cd{\U{39.7}}{40.2} & \cu{\U{0.067}}{0.002} &\cd{\U{43.1}}{13.4} \\
\midrule
\multicolumn{19}{c}{\cellcolor{groupband}\itshape Proprietary Chat LLMs} \\[1pt]
\midrule
Gemini-3 Flash~\cite{gemini_3_flash} & \cdL{\U{97.6}}{0.5} & \cdL{\U{98.8}}{0.7} & \cd{\Nb{5.710}}{0.726} &\cd{\Nb{81.1}}{8.9} & \cd{\U{53.2}}{14.2} & \cd{\Nb{76.2}}{6.8} & \cdL{\Nb{78.0}}{2.7} & \cd{\Nb{68.3}}{20.8} & \cz{\BU{93.3}} & \cdL{\Nb{69.7}}{4.5} & \cdL{\BU{74.9}}{4.4} & \cd{\Nb{51.5}}{11.2} & \cd{\Nb{33.5}}{20.2} & \cd{\U{38.0}}{22.0} & \cd{\Nb{49.5}}{32.8} & \cd{\Nb{59.5}}{20.4} & \cu{\U{0.069}}{0.000} &\cd{\Nb{47.9}}{8.6} \\
Claude Opus 4.6~\cite{claude_opus_4_6} & \cdL{\Nb{95.4}}{2.6} & \cdL{\Nb{98.3}}{1.3} &  \cd{\U{5.668}}{0.684} &\cdL{\BU{85.4}}{4.6} & \cd{\Nb{45.2}}{22.2} & \cz{\U{83.0}} & \cu{\U{81.8}}{1.2} & \cd{\U{78.1}}{11.0} & \cd{\Nb{52.9}}{40.3} & \cu{\BU{81.6}}{7.4} & \cd{\Nb{65.5}}{13.9} & \cd{\U{55.4}}{7.3} & \cd{\U{40.7}}{12.9} & \cd{\Nb{28.0}}{32.0} & \cdL{\BU{78.3}}{3.9} & \cd{\U{69.3}}{10.6} & \cdL{\Nb{0.072}}{0.003} &\cdL{\U{54.2}}{2.3} \\
\midrule
\multicolumn{19}{c}{\cellcolor{groupband}\itshape Open-source Chat LLMs} \\[1pt]
\midrule
Llama-3.3 70B~\cite{meta_llama33_70b_instruct} & \cd{\Nb{79.4}}{18.6} & \cd{\Nb{66.2}}{33.4} & \cd{\Nb{7.449}}{2.465} &\cd{\Nb{72.7}}{17.3} & \cd{\Nb{39.1}}{28.3} & \cd{\Nb{40.0}}{43.0} & \cd{\Nb{72.8}}{7.8} & \cd{\Nb{50.0}}{39.1} & \cd{\Nb{51.1}}{42.2} & \cd{\Nb{65.9}}{8.3} & \cd{\Nb{42.9}}{36.5} & \cd{\Nb{39.4}}{23.3} & \cd{\BU{46.7}}{7.0} & \cd{\Nb{38.0}}{22.0} & \cd{\Nb{32.9}}{49.4} & \cd{\Nb{63.8}}{16.1} & \cdL{\Nb{0.070}}{0.001} &\cd{\Nb{37.6}}{18.9} \\
Qwen2.5-Coder 32B~\cite{qwen25_coder_32b_instruct} & \cdL{\U{96.6}}{1.5} & \cd{\U{66.6}}{33.0} & \cd{\Nb{9.502}}{4.518} &\cd{\U{77.1}}{12.9} & \cd{\Nb{39.1}}{28.3} & \cd{\Nb{40.0}}{43.0} & \cd{\Nb{69.8}}{10.8} & \cd{\Nb{54.8}}{34.3} & \cd{\Nb{42.6}}{50.6} & \cu{\U{80.0}}{5.8} & \cd{\Nb{52.5}}{26.8} & \cd{\Nb{44.0}}{18.6} & \cd{\Nb{19.6}}{34.1} & \cd{\Nb{31.0}}{29.0} & \cd{\U{50.2}}{32.0} & \cd{\Nb{25.5}}{54.4} & \cdL{\Nb{0.117}}{0.048} &\cd{\Nb{38.4}}{18.1} \\
Kimi-k2.6~\cite{kimi_k2_6} & \cdL{\Nb{96.4}}{1.6} & \cd{\Nb{42.9}}{56.7}& \cd{\U{6.139}}{1.155}& \cd{\Nb{48.4}}{41.6} & \cd{\U{52.1}}{15.3} & \cz{\U{83.0}} & \cu{\BU{95.9}}{15.3} & \cd{\U{71.4}}{17.8} & \cd{\U{88.0}}{5.3} & \cdL{\Nb{70.5}}{3.7} & \cd{\U{70.4}}{9.0} & \cu{\BU{64.0}}{1.3} & \cd{\Nb{43.0}}{10.7} & \cd{\U{48.0}}{12.0} & \cd{\Nb{49.3}}{33.0} & \cdL{\U{79.1}}{0.8} & \cu{\BU{0.064}}{0.005} &\cu{\BU{57.2}}{0.7} \\
\midrule
\multicolumn{19}{c}{\cellcolor{groupband}\itshape Human reference} \\[1pt]
\midrule
\textit{Human Top} & \cellcolor{humanrow}\textit{98.0} & \cellcolor{humanrow}\textit{99.6} & 
\cellcolor{humanrow}\textit{4.98} & \cellcolor{humanrow}\textit{90.0} & \cellcolor{humanrow}\textit{67.4} & \cellcolor{humanrow}\textit{83.0} & \cellcolor{humanrow}\textit{80.6} & \cellcolor{humanrow}\textit{89.1} & \cellcolor{humanrow}\textit{93.3} & \cellcolor{humanrow}\textit{74.2} & \cellcolor{humanrow}\textit{79.3} & \cellcolor{humanrow}\textit{62.7} & \cellcolor{humanrow}\textit{53.7} & \cellcolor{humanrow}\textit{60.0} & \cellcolor{humanrow}\textit{82.2} & \cellcolor{humanrow}\textit{79.9} & 
\cellcolor{humanrow}\textit{0.069} & \cellcolor{humanrow}\textit{56.5} \\
\textit{Avg.\ human--LLM gap} & \cellcolor{gaprow}{\textcolor{gapblue}{$\downarrow6.6$}} & \cellcolor{gaprow}{\textcolor{gapblue}{$\downarrow18.8$}} & \cellcolor{gaprow}{\textcolor{gapblue}{$\downarrow1.333$}} & \cellcolor{gaprow}{\textcolor{gapblue}{$\downarrow18.0$}} & \cellcolor{gaprow}{\textcolor{gapblue}{$\downarrow17.1$}} & \cellcolor{gaprow}{\textcolor{gapblue}{$\downarrow14.6$}} & \cellcolor{gaprow}{\textcolor{gapblue}{$\downarrow1.3$}} & \cellcolor{gaprow}{\textcolor{gapblue}{$\downarrow19.1$}} & \cellcolor{gaprow}{\textcolor{gapblue}{$\downarrow24.8$}} & \cellcolor{gaprow}{\textcolor{gapblue}{$\downarrow4.0$}} & \cellcolor{gaprow}{\textcolor{gapblue}{$\downarrow17.0$}} & \cellcolor{gaprow}{\textcolor{gapblue}{$\downarrow13.5$}} & \cellcolor{gaprow}{\textcolor{gapblue}{$\downarrow19.4$}} & \cellcolor{gaprow}{\textcolor{gapblue}{$\downarrow22.4$}} & \cellcolor{gaprow}{\textcolor{gapblue}{$\downarrow23.0$}} & \cellcolor{gaprow}{\textcolor{gapblue}{$\downarrow17.1$}} & \cellcolor{gaprow}{\textcolor{gapblue}{$\downarrow0.008$}}& \cellcolor{gaprow}{\textcolor{gapblue}{$\downarrow9.8$}} \\
\bottomrule
\label{tab:gnn-llm-leaderboard}
\end{tabular}
\end{adjustbox}
\end{table*}

\begin{figure*}[!t]
  \centering
  \includegraphics[width=\textwidth]{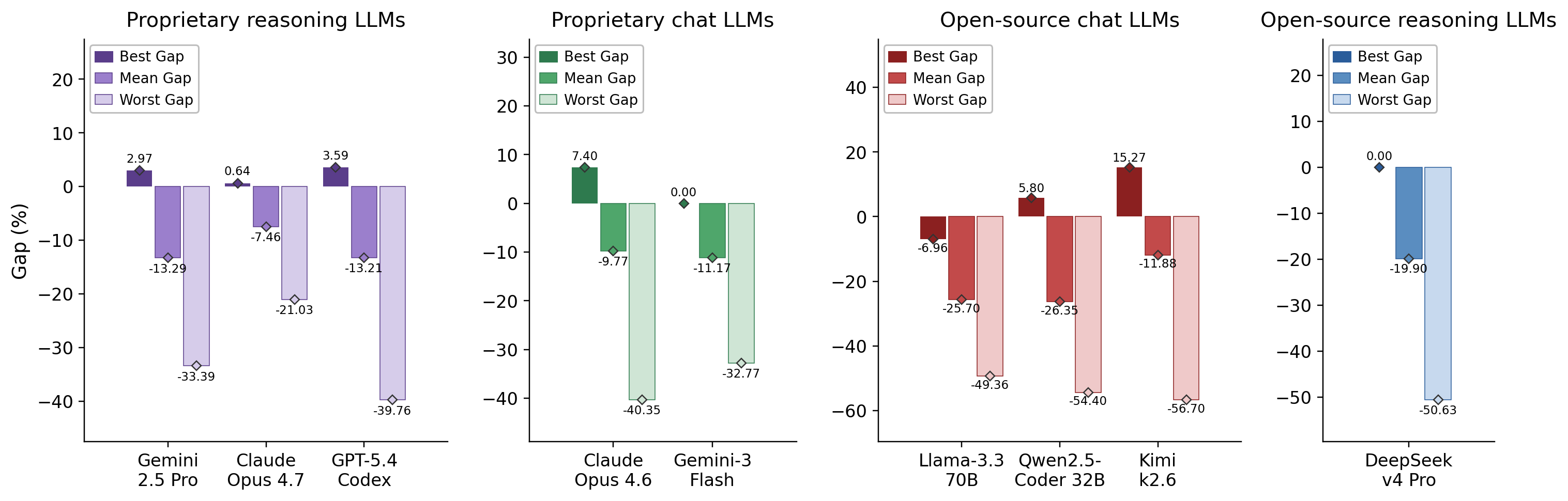}
\caption{Best, Mean, and Worst Gap performance for leading LLMs on the classification tasks of the \textsc{GNN-CB} benchmark, grouped by model family. \textbf{Best Gap} denotes the best performance margin across the 16 classification competitions compared to the human top performance. \textbf{Mean Gap} is the average performance difference relative to the human top performance across all classification competitions. \textbf{Worst Gap} represents the most negative performance difference relative to the human top performance.}
  \label{fig:llm-leaderboard}
\end{figure*}

\subsection{Human Performance Results}
\label{subsec:human_results_anchor}

Table~\ref{tab:human} and Figure~\ref{fig:human_results} together summarize the performance of human participants across all 18 competitions. Overall, human performance exhibits variability across tasks, reflecting differences in task difficulty, graph structure, and domain complexity. Easy classification tasks (e.g., C03) achieve high scores with low variance, indicating that standard GNN pipelines are implemented reliably when task specifications are clear. In contrast, harder competitions (e.g., C02, C04, C18) show lower performance and larger gaps between best and worst submissions, reflecting higher sensitivity to architectural and optimization choices. Several medium-difficulty tasks (e.g., C09, C10) also exhibit notable variability, suggesting that engineering and training decisions remain critical even for relatively standard formulations. The two regression tasks (C14 and C17) display similar behaviour, with larger relative performance gaps and less stable convergence compared to most classification settings. These observations suggest that human participants perform consistently on conventional GNN problems but encounter increasing difficulty as tasks require more advanced graph reasoning, domain-specific adaptation, and complex optimization strategies.

\subsection{LLM Evaluation Results}
\label{subsec:llm_results_anchor}


Table~\ref{tab:gnn-llm-leaderboard} reports the per-competition performance of the nine evaluated LLMs on all 18 \textsc{GNN-CB} tasks. The most important finding is that LLMs consistently fall short of Human Top performance across the benchmark under a zero-shot plan-then-code prompting protocol with a bounded execute-and-repair loop. Concretely, across all 9 evaluated models and 18 competition tasks (162 model–task pairs in total), LLMs surpass or tie the Human Top score in only 19 cases. In other words, in more than 88\% of model--task pairs, the evaluated LLM does not reach Human Top performance. Among models achieving Human Top-level performance, we observe no clear model-family advantage, and the Human-Top level performance counts are roughly proportional to the number of models in each family. In addition, the average LLM performance exceeds the average human performance in only 6 of 16 competitions (37.5\%), as shown in Figure~\ref{fig:avg-all}, meaning that average human performance remains higher on most competitions under this protocol.

To further assess LLM performance fairly, we analyze the gap to Human Top performance across 16 classification competitions. The evaluated open-source models exhibit larger average gaps and greater cross-task variability under this protocol. This trend is observed for both reasoning and chat models. DeepSeek-v4 Pro underperforms proprietary reasoning counterparts in mean gap, while open-source chat models (Llama-3.3 70B, Qwen2.5-Coder 32B) show larger gaps than proprietary chat models (Claude Opus 4.6, Gemini-3 Flash). An exception is Kimi-k2.6, which approaches the consistency of proprietary chat models. Within the proprietary models evaluated, neither reasoning nor chat models show a consistent benchmark-wide advantage, as also shown in Figure~\ref{fig:llm-leaderboard}. In addition, LLMs exhibit larger performance gaps on Hard competitions than on Easy and Medium competitions, suggesting that the observed human--LLM gap increases with task difficulty under this protocol.

We further examined severe underperformance in classification competitions, defined as scoring at least 30 percentage points below the best-performing LLM on the same task. Manual inspection identified 20 such cases, with Llama-3.3 70B and Qwen2.5-Coder 32B accounting for six and five cases, respectively. The error analysis showed that these failures were driven primarily by incorrect graph construction (8/20) and feature processing or utilization (7/20), rather than by minor coding issues or GNN architecture choice. The remaining cases involved indexing or data-reading errors (3/20) and unsuitable model selection (2/20). This suggests that severe degradation mainly arises when models fail to translate the input data into an appropriate graph representation.Further results are presented in Appendix~\ref{app:llm-perf}.

\section{Discussion}
\label{sec:analysis}

The findings in Table~\ref{tab:gnn-llm-leaderboard} and Figure~\ref{fig:llm-leaderboard} reveal the core limitations of LLMs as autonomous GNN engineers under the evaluated setting. All evaluated models show substantial variance across the 18 competitions, indicating limited transfer of GNN construction and reasoning strategies across heterogeneous settings.Proprietary models demonstrate better average gaps than their open-source counterparts, yet no meaningful difference emerges between reasoning and chat LLMs within either tier—suggesting that chain-of-thought reasoning capabilities do not confer a systematic advantage on GNN-specific tasks under our zero-shot, bounded-repair protocol. In addition, performance gaps widen on Hard tier competitions, which shows the inability of LLMs dealing with complex GNN-based tasks. These results suggest that task-aware model routing or ensembling may be more effective than scaling a single model. Competition-style evaluation also provides a discriminative measure of practical GNN coding ability and highlights the risks of deploying LLM-generated solutions without expert oversight.

\section{Conclusion}
\label{sec:conclusion}
We introduced \textsc{GNN-CB}, the first \emph{competition-based} benchmark for evaluating humans and LLMs on GNN coding tasks. It comprises 18 human-designed competitions spanning diverse graph categories, domains, and difficulty tiers, evaluated through a unified hidden-test pipeline. Three complementary protocols govern competition design, controlled human evaluation, and frozen-prompt LLM evaluation with bounded execute-and-repair iterations across API-only and agent-based settings. Under this protocol, current LLMs fall short of expert performance on most tasks and exhibit inconsistent performance across heterogeneous graph-learning problems. As models and evaluation methods continue to improve, the competition-based design of \textsc{GNN-CB} offers a challenging, extensible, and practically grounded testbed that can continuously distinguish advances in autonomous GNN engineering. Its public benchmark and dynamic leaderboards support future evaluation, comparison, and hands-on GNN coding practice.

\section*{Limitations}

Due to API and compute-budget constraints, our LLM evaluation is limited to nine leading proprietary and open-source models across two execution modes (API-only and agent-based), and our human evaluation involves 13--17 expert volunteers per competition. The model and task selection was guided by a balance between inclusiveness and feasibility, aiming to cover diverse access types (proprietary vs.\ open-source), and model families (Claude, Gemini, GPT, Llama, Qwen, Kimi, DeepSeek) while maintaining reproducibility and accessibility for research groups with limited compute resources. We believe this setup provides a meaningful and representative evaluation of contemporary LLM capabilities on end-to-end GNN coding tasks, without requiring the extensive infrastructure typically available only to large industrial laboratories. Accordingly, our results should be interpreted as a broad and reproducible evaluation of contemporary LLM performance under the evaluated bounded-repair and agent-based protocols, rather than as an estimate of the maximum attainable performance of each model. Larger inference budgets, repeated sampling, and more advanced model-specific prompting or agentic pipelines may further improve performance and should be investigated in future work.

A further limitation concerns the enforceability of the human-track constraint. Although the protocol restricts LLM use to low-level utility assistance and requires detailed reproducibility logs, local use of external assistance cannot be perfectly instrumented. Therefore, this constraint should be interpreted as practically auditable rather than fully enforceable.

While \textsc{GNN-CB} covers a broad range of node-, edge-, and graph-level problems, certain emerging settings such as graph generation, quantum graph modelling, and self-supervised or reinforcement-learning paradigms are not included in the current release. The use of task-specific evaluation metrics (macro-$F_1$, accuracy, AUC, MAE, and NMAE) reflects the heterogeneity of real-world graph problems and limits direct cross-task comparison. To mitigate this, we report per-task gap-to-human-top scores and separate classification and regression results in our visualizations, particularly since the benchmark contains only two regression competitions. Although the two regression competitions broaden the task and metric diversity of \textsc{GNN-CB}, their limited number does not support strong general conclusions about LLM performance on graph-regression problems. Expanding both the number and diversity of regression competitions is therefore an important direction for future releases. Finally, because LLM capabilities evolve rapidly, the reported results represent model snapshots available as of May~2026, although the qualitative patterns we observe---difficulty stratification, the capability-versus-consistency gap, and family-level differences---are expected to remain informative as models continue to advance. These limitations can be viewed as opportunities for future extensions of \textsc{GNN-CB} toward broader graph-learning paradigms, larger model panels, and continuously updated evaluations.

\section*{Future Directions for Extended LLM Evaluation}
\label{app:future_work}

The present study establishes \textsc{GNN-CB} as a zero-shot, frozen-prompt benchmark, and its infrastructure enables several deeper analyses that we did not pursue here due to computational and budgetary constraints. On the modelling side, future work can measure how performance scales beyond the current zero-shot protocol through few-shot prompting, retrieval-augmented generation, and task-specific fine-tuning, and can characterise stochastic variability via repeated sampling, confidence intervals, and pass@$k$-style metrics rather than the single greedy run reported here. Because every submission is executed and validated, the benchmark also supports a fine-grained error-mode taxonomy---incorrect graph construction, invalid message-passing architectures, preprocessing data leakage, mishandled class imbalance, and malformed submissions---that would reveal \emph{where} and \emph{why} LLMs break down beyond aggregate scores.

On the evaluation side, the logged execution and repair traces enable a detailed token-economy and cost--latency analysis, particularly contrasting the per-call budget of API-only runs with the multi-step session budget of agent-based runs, alongside an analysis of agentic depth (tool-call counts, repair-loop depth, and the contribution of extended chain-of-thought). Rigour can be further strengthened through paired significance testing, bootstrap confidence intervals, and rank-correlation analysis of the human--LLM ordering. Finally, as a living benchmark with automated infrastructure and dynamic leaderboards, \textsc{GNN-CB} can be extended to broader paradigms (graph generation, self-supervised, temporal, and reinforcement-learning formulations) and used longitudinally to track whether the capability--consistency gap narrows as new models are released.
\section*{Ethics Statement}

\paragraph{Participant Consent:}
All human participants provided informed consent for the publication
of their performance results. Participants were expert volunteers
who co-designed the benchmark; no monetary compensation was provided.
Only aggregate counts and per-task scores are reported (Table~\ref{tab:human}); no personally identifying information about participants is released.

\paragraph{Participant Role Disclosure:}
Each of the 18 benchmark competitions was designed and curated by a single coauthor. To avoid any designer advantage, no coauthor participated in the competition they personally designed. Instead, each coauthor solved a subset (approximately 14) of the remaining 17 competitions. Consequently, every human submission contributing to the Human Top, mean, and standard deviation reported for a given competition was produced by a participant with no design or curation role in that specific competition.

\paragraph{Personally Identifying Information and Sensitive Content:}
All datasets are used in their publicly released forms. Sensitive
biomedical datasets such as OASIS-3 ~\cite{lamontagne2019oasis} and ABIDE ~\cite{graph4ASDdataset}
follow their original de-identification protocols and data-use
agreements; only aggregate per-task scores are reported. The GitHub
Social Network dataset \citep{rozemberczki2019multiscale} contains
public GitHub usernames that are already openly available; the
usernames are not the target of prediction and no additional
de-anonymization is performed. The Elliptic dataset
\citep{weber2019anti} uses pseudo-anonymous transaction identifiers
with no link to real identities. None of the datasets contain
free-form user-generated text in their released versions, so no
offensive-content screening was required.

\paragraph{Licenses:}
All datasets are used under their original public licenses
(MoleculeNet, Cora, OGB, OASIS-3 under data-use agreement, Elliptic,
GossipCop, GitHub Social Network, etc.). LLMs are accessed under
the providers' API terms of service. GNN-CB itself, including the
benchmark repositories, evaluation pipeline, and leaderboard
infrastructure, will be released under the MIT license.

\paragraph{Potential Risks:}
GNN-CB evaluates LLM-generated GNN code that, if deployed in
safety-critical domains such as medical diagnosis, financial-fraud
detection, or neuroimaging without expert verification, could
amplify model failures or propagate dataset biases. We release the
benchmark for controlled research and hands-on GNN coding practice, and discourage
direct deployment of LLM-generated solutions without expert review.

\paragraph{Use of AI Assistants.}
AI assistants were used solely for editorial and coding support 
during paper preparation, including grammar correction, LaTeX 
formatting, table and figure layout, and minor stylistic 
polishing. AI assistants were not used to generate experimental 
results, write benchmark code, or produce scientific claims. 
All scientific content---research questions, benchmark design, 
dataset curation, experimental design, analyses, and 
conclusions---was conceived and verified by the human authors. 
The AI assistants used for paper preparation are distinct from 
the LLMs evaluated in the benchmark.

\bibliographystyle{unsrtnat}
\bibliography{references}

\appendix

\section{Competition Specifications}
\label{app:specs}

This appendix provides detailed descriptions of the competitions included in \textsc{GNN-CB}, each following a standardized structure (task definition, dataset statistics, evaluation metric, and submission protocol) and a unified evaluation pipeline based on hidden test sets, secure submission handling, and automated GitHub Actions scoring to ensure reproducibility and fairness.

To mitigate the potential effects of data contamination from widely used public datasets, benchmark competitions are not restricted to canonical dataset configurations. Instead, most competitions in \textsc{GNN-CB} apply controlled modifications before constructing the benchmark task. These modifications fall into three main categories:

\begin{itemize}
    \item \textbf{Feature and data modifications.} Some competitions modify the original data to increase task difficulty and reduce the transferability of memorized solutions. For example, \textbf{C12 (Cora-GNN)} injects Gaussian noise into node features, \textbf{C10 (NetLinkArena)} applies feature permutation and noise injection, and \textbf{C13 (GraphFMD)} combines feature perturbation with a temporal split and adjusted class balance.

    \item \textbf{Distribution changes and new data splits.} Several competitions construct new training and testing settings that differ from the original benchmark. For example, \textbf{C05 (GNN-BACE)} introduces severe class imbalance, applies scaffold splitting, and removes pre-computed descriptors. \textbf{C16 (Graph4ASD)} introduces a cross-site distribution shift, while \textbf{C14 (BrainAge)} uses new age-stratified subject splits and a restricted set of imaging features.

    \item \textbf{Converting non-graph datasets into graph-learning tasks.} Some competitions reformulate the original dataset into a graph-learning problem. For example, \textbf{C11 (PARK-GNN)} transforms the original tabular Parkinson's dataset into a graph using similarity-based edges. Likewise, \textbf{C06 (DiaGraph)} and \textbf{C07 (GRAPE)} convert non-graph data into graph representations. Although the underlying data remain unchanged, the resulting graph-learning tasks have no direct public reference solutions.
\end{itemize}

Across all competitions, the goal is the same: to reduce the direct transferability of memorized solutions from LLM pretraining while creating more realistic and challenging graph-learning tasks. The exact modification depends on the characteristics of each dataset and application.

The benchmark spans a diverse set of graph problem categories, including \textcolor{catHomo}{homogeneous}, \textcolor{catHetero}{heterogeneous}, \textcolor{catTemporal}{temporal}, \textcolor{catSpatial}{spatial}, \textcolor{catHier}{hierarchical}, \textcolor{catMol}{molecular}, and \textcolor{catGen}{generalization} graphs. To fomally define, \textcolor{catHomo}{Homogeneous} : Graph with uniform node and edge types, \textcolor{catHetero}{Heterogeneous} : Graph with  multiple node and edge type, \textcolor{catTemporal}{Temporal} : Graph changes over time, \textcolor{catSpatial}{Spatial}: Graph is embedded in a spatial domain where nodes and edges reflect physical or geometric proximity, \textcolor{catHier}{Hierarchical}: Graph with multi-level structure among nodes, \textcolor{catMol}{Molecular}: Graph represents molecular structures, where nodes are atoms and edges are chemical bonds, \textcolor{catGen}{Generalization}: Test graph is drawn from distributions different from the training graph.

We acknowledge that some categories may overlap; however, we assign each competition to its most distinct category. For example, a molecular graph is also homogeneous, but it is labeled as \textcolor{catMol}{Molecular} to distinguish it from other types.

Across all competitions, a unified submission and evaluation protocol is enforced to ensure fairness, reproducibility, and consistent leaderboard construction. Each participant is allowed a single final submission per competition version, and only the latest valid submission is considered for evaluation. Submissions that are not properly encrypted, fail decryption, or do not conform to the required format are automatically rejected and excluded from scoring. Leaderboard ranking follows a Kaggle-style tie-breaking rule, where identical scores are assigned the same rank, and no additional ordering is enforced between tied entries.
\subsection*{C01: PROVEN-GNN}

\textbf{PROVEN-GNN} (\textbf{PRO}gram \textbf{V}ulnerability \textbf{E}xamination \textbf{N}etwork using \textbf{G}raph \textbf{N}eural \textbf{N}etworks) is a graph classification task on \textcolor{catHetero}{heterogeneous} graph representations of code functions. Each source code function is represented as a graph $G_k=(V_k,E_k)$, where nodes denote code fragments (methods, variable declarations, etc.) and edges represent the relationships between them (AST, CFG, etc.). The objective is to detect vulnerable code functions based on these representations.

The dataset consists of $3109$ functions from the \textit{DiverseVul} dataset \cite{PROVEN-GNN}, with vulnerable functions representing $29\%$ of the data. The graph representations were generated using the tool \textit{Joern}. It allows generating code property graphs that capture syntactic properties, control flows, and data dependencies. Each node is represented with $512$ features, while edge features are represented using a one-hot encoding of the edge type ($12$ features).

Participants receive training and test graph files containing node embeddings, edge indices, and edge features. The labels ($1$ for vulnerable and $0$ for non-vulnerable) are provided only for the training graphs. Participants are then required to generate predictions for the test graphs and submit them in an encrypted form using the provided public key. An automated pipeline then validates, decrypts, scores, and updates the leaderboard, evaluating each submission using the macro F1-score.

\subsection*{C02:CGCC}

The \textbf{CGCC} (\textbf{C}ity \textbf{G}raph \textbf{C}lassification \textbf{C}hallenge) is a graph-level classification task defined over real-world urban street networks. Each sample is a city represented as a \textcolor{catHomo}{homogeneous} graph $G_k=(V_k,E_k)$, wherenodes denote road intersections and edges denote the road segments connecting them. Given a collection of such graphs $\mathcal{G}=\{G_k\}$, the objective is to learn a mapping $f(G_k)\rightarrow y_k$ that assigns each city to one of three layout classes: organic (0), grid (1), or hybrid (2). 

The dataset comprises street networks for 120 cities distributed across North America, South America, Europe, Asia, Africa, Oceania, the Middle East, and Central Asia, extracted from OpenStreetMap via OSMnx within a 500\,m buffer around each city centre and restricted to driveable roads. The class distribution reflects naturally occurring urban patterns and is moderately imbalanced, with 37 grid, 31 organic, and 52 hybrid cities. Each graph retains node coordinates, edge geometries and lengths, and graph-level attributes such as the city name and layout type; for the evaluation split, the layout label is withheld. The corpus is partitioned $70/30$ into training and test sets with stratification by class.

Participants are provided with serialized city graphs and a label table that associates each training graph with its target class. No engineered urban-form descriptors (e.g., circuity, intersection density, orientation entropy) are provided. Participants are required to generate a prediction file mapping each test city to its predicted class, encrypt it locally using the provided public key, and submit encrpted file through a pull-request-based workflow. An automated pipeline then validates, decrypts, and scores the submissions, updating the leaderboard. Each submission is evaluated using the macro F1-score.

\subsection*{C03: Real Or Fake!}

The \textbf{Real or Fake} challenge is a graph classification task based on Twitter news propagation data from the \textit{GossipCop} dataset \citep{RealOrFakeDataset1,RealOrFakeDataset2}. The objective is to train a Graph Neural Network (GNN) that predicts whether a news article is real or fake by leveraging both node features and the underlying propagation structure. Given a collection of \textcolor{catHier}{hierarchical} graphs $\mathcal{G} = \{G_k\}$, where each graph $G_k = (V_k, E_k)$ represents the spread of a single news article, the task is to predict the label $y_k$ for each graph :$f(G_k) \rightarrow y_k$ where $y_k \in \{0,1\}$, with $1$ indicating verified news and $0$ indicating fake news.

In this setting, nodes correspond to either the root news article or users who interacted with it, while edges capture interactions such as retweets. Each node is associated with rich features, including textual embeddings (e.g., BERT and spaCy) and user profile attributes. The dataset is split into 3,826 training graphs, 546 validation graphs, and 1,092 test graphs.

Participants are provided with the graph structure, node features, and data splits distributed across multiple files. They are then expected to generate predictions for the test graphs, encrypt them locally using the provided public key and and submit through a pull-request. An automated pipeline subsequently validates, decrypts, and evaluates the submissions, updating the leaderboard accordingly. Each submission is assessed using the accuracy score.

\subsection*{C04: GLIMPS-GNN}

\textbf{The GLIMPS-GNN} (\textbf{G}raph-based \textbf{L}iquid-biopsy \textbf{I}nductive \textbf{M}odeling for \textbf{P}reeclamp\textbf{S}ia) challenge is an inductive node-classification task on a biological graph. Given $G = (V, E)$ with feature matrix $\mathbf{X} \in \mathbb{R}^{|V| \times d}$, the goal is to learn a mapping $f(v_i) \rightarrow y_i$  where each node $v_i \in \mathcal{V}$ is assigned a label $y_i \in \{0,1\}$ corresponding to Control (0) and Preeclampsia (1). The training subgraph comprises cell-free RNA (cfRNA) from maternal plasma, while the inductive test set contains unseen placental RNA-seq samples. Models must therefore generalize across a substantial domain shift between liquid-biopsy and tissue transcriptomics, making this competition fall under the \textcolor{catGen}{generalization} category.

Data are sourced from GEO: maternal plasma cfRNA (GSE192902) and placental RNA-seq (GSE234729). Nodes are individual samples; edges encode similarity and ancestry. The graph contains 320 nodes and 3200 edges. Node features are harmonized gene-expression profiles. The training set has 209 cfRNA samples; the test set has 111 placenta samples with moderately imbalanced classes.

Participants are provided with the graph structure, node features, and data splits distributed across multiple files. They are then expected to generate predictions for the test graphs and encrypt them locally using the provided public key. After submission, an automated GitHub Actions pipeline subsequently validates, decrypts, and evaluates the submissions, updating the leaderboard accordingly. Each submission is assessed using the macro-averaged F1-score.

\subsection*{C05: GNN\_BACE} 

\textbf{GNN\_BACE} (BACE Inhibition Prediction) is a graph-level binary classification task whose primary objective is to predict whether a given molecule acts as an inhibitor of the $\beta$-secretase 1 (BACE-1) enzyme, a critical target in Alzheimer’s disease research. Formally, the objective is to learn a mapping $f(G_k)\rightarrow y_k$ for each molecule represented as  \textcolor{catMol}{molecular} graphs $G_k = (V_k, E_k)$. Molecules are represented as undirected graphs $G_k = (V_k, E_k)$, where nodes correspond to atoms and edges correspond to chemical bonds. Each node is described by an 8-dimensional one-hot atom-type feature vector, while edges include simple bond-level attributes.  Label is $1$ for active inhibitors and inactive compounds as $0$.

Participants are provided only with atom features and connectivity information, without any pre-computed molecular descriptors (e.g., MolLogP, Polar Surface Area). The dataset for this challenge is derived from the BACE dataset in the widely used MoleculeNet benchmark suite \citep{GNN-BACE}. Participants are then required to generate predictions for the test graphs and encrypt them locally using the provided public key. After submitting encrypted file using pull-request, an automated GitHub Actions pipeline subsequently validates, decrypts, and evaluates the submissions, updating the leaderboard accordingly. Each submission is assessed using the Macro $F_1$-Score.

\subsection*{C06: DiaGraph}

The \textbf{DiaGraph} (Robust Diabetes Prediction on Noisy Graphs) is a transductive node-level classification task where each sample corresponds to a patient represented as a node in a \textcolor{catHomo}{homogeneous} graph $G=(V,E)$. Nodes denote patients, while edges encode similarity relationships based on latent clinical affinities. The objective is to learn a mapping $f(v_i)\rightarrow y_i$ that predicts whether a patient is diabetic ($1$) or non-diabetic ($0$).

Participants are provided only with node features and the adjacency matrix for all nodes, while labels are available only for the training nodes. For evaluation, participants generate a prediction file containing binary labels for all nodes in the hidden test set. The predictions are encrypted locally using the provided public key prior to submission, and only encrypted files are submitted to the repository. An automated evaluation pipeline decrypts the submissions, validates their format, computes the official score against hidden test labels, and updates the public leaderboard. Each submission is evaluated using the macro F1-score.

\subsection*{C07: GRAPE }

The \textbf{GRAPE} (\textbf{G}raph \textbf{R}etinal \textbf{A}nalysis for \textbf{P}rediction and \textbf{E}valuation) challenge is a graph-level classification benchmark designed for diabetic retinopathy (DR) detection from retinal vessel graphs. Each sample corresponds to a retinal vascular network represented as a \textcolor{catSpatial}{spatial} graph $G_k=(V_k,E_k)$, where nodes denote vessel bifurcation or intersection points and edges represent vessel segments connecting them. Given a collection of retinal graphs $\mathcal{G}=\{G_k\}$, the objective is to learn a mapping $f(G_k)\rightarrow y_k$ that predicts whether the retinal image corresponds to a healthy subject ($0$) or a diabetic retinopathy patient ($1$).

The benchmark is constructed from retinal vessel graphs extracted from three publicly available retinal imaging datasets: DRIVE, STARE, and HRF. In total, the dataset contains 70 graphs, comprising 48 healthy and 22 diabetic retinopathy cases. Specifically, the DRIVE subset contributes 20 graphs (17 healthy, 3 DR), STARE contributes 20 graphs (16 healthy, 4 DR), and HRF contributes 30 graphs (15 healthy, 15 DR). The dataset is partitioned into 55 training graphs and 15 hidden test graphs using stratified sampling to preserve class proportions across splits. Each retinal graph preserves the vascular connectivity structure derived from vessel segmentation outputs.

Participants are provided with node features and the adjacency matrix for all graphs, while labels are available only for the training graphs. Participants are expected to generate a prediction file containing binary labels for all graphs in the hidden test set. The predictions are encrypted locally using the provided public key prior to submission, and only encrypted files are submitted to the repository. An automated GitHub Actions workflow securely injects the private decryption keys, decrypts the submissions, validates their format, computes the official score against the hidden test labels, and updates the public leaderboard. Each submission is evaluated using the macro F1-score.

\subsection*{C08: Liar Nodes }

The \textbf{Liar Nodes} challenge is formulated as a transductive node classification problem over a \textcolor{catHomo}{homogenous} graph derived from single-cell cancer transcriptomics data. Each node represents an individual cell, while graph edges encode contextual or biological similarity relationships between cells within the tumor microenvironment. Given a graph $G=(V,E)$, the objective is to learn a mapping $f(v_i)\rightarrow y_i$ that predicts whether a cell is cancerous or non-cancerous.

The underlying data originate from the \textbf{Cancer Single-cell Expression Map (CancerSCEM)} resource \citep{LiarNodes}, which contains large-scale single-cell RNA sequencing data across multiple cancer types. The original dataset provides UMI count matrices, cell-type annotations, functional molecular descriptors, and cell-cell interaction information. For the purposes of this benchmark, the data are transformed into graph learning instances suitable for GNN training and evaluation. Node embeddings are intentionally altered and partially corrupted in order to create contradictions between intrinsic node representations and graph context.

Participants are provided with graph-structured data derived from the CancerSCEM resource, together with training labels for supervised learning. Final predictions are submitted via a designated Google Form for evaluation and leaderboard ranking. An automated evaluation pipeline computes accuracy on hidden test labels and updates the public leaderboard accordingly.

\subsection*{C09: GraphFMD}
The \textbf{GraphFMD} (\textbf{Graph}-based \textbf{F}inancial \textbf{M}isconduct \textbf{D}etection) challenge is a temporal inductive node classification task defined over the Bitcoin transaction network. Each sample is a transaction represented as a node $v_i$ in a directed \textcolor{catTemporal}{temporal} graph $G=(V,E)$, where nodes denote individual Bitcoin transactions and edges denote the flow of Bitcoin between them. The objective is to learn a mapping $f(v_i) \rightarrow y_i$ that assigns each test-time transaction to one of two classes: illicit (1) or licit (2). 

The dataset is derived from the Elliptic Bitcoin transaction graph~\cite{weber2019anti} and is partitioned temporally: training nodes span time steps 1--34, comprising 16{,}658 labelled transactions, while test nodes are drawn from time steps 35 and beyond, comprising 8{,}896 transactions. The graph contains approximately 5 million directed edges. Each node is associated with a 165-dimensional feature vector consisting of 94 local transaction features (e.g., transaction amounts, fees, and input/output counts) and 71 aggregated neighbourhood features derived from one-hop neighbours; Gaussian noise has been injected into these features to simulate real-world data degradation. The dataset exhibits strong class imbalance, with illicit transactions constituting only 1--3\% of all nodes.

Participants are provided with node feature files and an edge list for the full graph, together with ground-truth labels for the training split. No hand-crafted forensic descriptors (e.g., mixing patterns, chain-length heuristics, or address-clustering signals) are supplied. Participants are expected to generate a prediction table mapping each test transaction identifier to its predicted class ${1, 2}$. Submissions are encrypted and uploaded via a secure Google Form, which triggers a GitHub Actions workflow that validates the submission, computes the performance against the hidden label file, and updates the public leaderboard automatically. Performance is evaluated on the hidden test split using the macro F1-score.

\subsection*{C10: NetLinkArena}

The \textbf{NetLinkArena} (Link Prediction Challenge) is a link prediction task defined on a real-world scientific citation network. Each sample corresponds to a pair of scientific papers represented within a \textcolor{catHomo}{homogenous} citation graph $G=(V,E)$, where nodes denote research papers and edges represent citation relationships between them.  Given a partially observed graph and a set of candidate node pairs $(u,v)$, the objective is to learn a mapping $f(u,v) \rightarrow y$ that predicts whether a citation link exists between the two papers, where $y \in \{0,1\}$, with $1$ indicating the presence of a link and $0$ indicating no link.

The dataset is constructed from the original \textbf{CiteSeer} dataset \cite{sen2008collective}, in which additional perturbations and synthetic noise were introduced to generate a more challenging and realistic link prediction benchmark. The graph contains 3,327 scientific papers, each represented by a 2,742-dimensional feature vector that captures the semantic and structural properties of each document. Citation relationships are provided as source-target node pairs, where positive edges correspond to observed citations and negative edges represent non-citation pairs sampled from the graph. The dataset is partitioned into training, validation, and hidden test splits for supervised evaluation.

Participants are provided with the node feature matrix, labeled source–target node pairs for training and validation, and unlabeled node pairs for inference. Participants are expected to generate a file mapping each test node pair to a predicted probability score indicating the likelihood of a citation link. Submissions are entered through a Google Sheets-based workflow, where each team records predictions in a designated submission sheet. A GitHub Actions workflow then validates the submission format and updates the leaderboard automatically. Performance is evaluated on the hidden test split using the AUC score.

\subsection*{C11: PARK-GNN}

The \textbf{PARK-GNN} Challenge (\textbf{P}arkinson's \textbf{A}coustic \textbf{R}epresentation \& \textbf{K}nowledge with \textbf{G}raph \textbf{N}eural \textbf{N}etworks) is a transductive node classification task defined over a \textcolor{catHomo}{homogenous} graph constructed from acoustic voice recordings. Each node $v_i$ in the graph $G=(V,E)$ represents a single voice recording, characterized by a 22-dimensional feature vector of acoustic measurements including vocal fundamental frequency statistics, jitter and shimmer variations, harmonics-to-noise ratios, and nonlinear dynamical complexity features. Edges are formed by a $k$-nearest-neighbor criterion ($k=5$) over acoustic similarity in feature space, supplemented by subject-level connections that link recordings from the same individual. The objective is to learn a mapping $f(v_i)\rightarrow y_i\in\{0,1\}$ that assigns each voice recording to either a healthy control (0) or a Parkinson's Disease patient (1).

The dataset is derived from the UCI Parkinson's Dataset \citep{PARK-GNN}, comprising 195 voice recordings collected from 31 subjects, of whom 23 were diagnosed with Parkinson's Disease and 8 were healthy controls. The corpus is partitioned into a training set of 156 nodes (approximately 80\%) and a hidden test set of 39 nodes (approximately 20\%). Training labels are provided alongside the full graph structure; test labels are withheld and used exclusively within the automated evaluation pipeline.

Participants are provided with the serialized graph structure, node feature matrices, and a label table mapping each training node to its ground-truth class. No engineered biomarker composites or pre-computed diagnostic descriptors are supplied.Participants are expected to generate a prediction table mapping each test node identifier to its predicted class. Submissions are encrypted locally using the provided public-key scheme and uploaded through a pull-request-based workflow. An automated GitHub Actions pipeline then validates, decrypts, scores, and updates the public leaderboard. Performance is evaluated on the hidden test split using the macro-averaged F1-score.

\subsection*{C12: GNN-CoRA}

The \textbf{GNN-CoRA} (Cora Node Classification Challenge) competition is a node classification task defined over a \textcolor{catHomo}{homogenous} citation-network graph derived from the \textit{Cora} benchmark dataset \citep{sen2008collective}. The objective is to train a GNN that predicts the research-topic label of each publication node from both its textual feature representation and the surrounding citation structure. Formally, given a graph $G=(V,E)$ in which nodes correspond to publications and edges denote citation relationships, the task is to learn a mapping $f(v_i)\rightarrow y_i$ that assigns each node $v_i$ to one of seven research-topic classes.

The dataset consists of 2,708 publication nodes connected by 5,429 citation edges. Each node is associated with a high-dimensional sparse bag-of-words representation extracted from the publication text, and the graph structure captures citation links between papers. The benchmark follows the standard citation-network split, containing 140 labelled training nodes, 500 validation nodes, and 1,000 hidden test nodes. To increase task difficulty and evaluate robustness, Gaussian noise is injected into the node-feature representations during benchmark construction.

Participants are provided with graph connectivity information, node feature matrices, training and validation label tables, and identifiers for the hidden test nodes. The ground-truth labels for the test split remain concealed and are used exclusively within the automated evaluation pipeline.  Participants are expected to generate a prediction table mapping each hidden test node to one of the seven target classes. Submissions are encrypted locally using the benchmark’s public-key encryption scheme and submitted through a pull-request-based workflow. An automated GitHub Actions evaluation pipeline then validates, decrypts, scores, and updates the leaderboard accordingly. Performance is evaluated on the hidden test set using classification accuracy.

\subsection*{C13 : GTA}

The \textbf{GTA} (\textbf{G}raph \textbf{T}opology \textbf{Ablation}) challenge is formulated as a supervised binary graph classification problem over molecular graphs derived from the \textbf{MUTAG} dataset, a widely used benchmark in chemical informatics. Each graph $G_k=(V_k,E_k)$ represents a chemical compound in which nodes correspond to atoms and edges correspond to chemical bonds. The objective is to learn a mapping $f(G_k)\rightarrow y_k$ that predicts whether a molecule is mutagenic or non-mutagenic with respect to \textit{Salmonella typhimurium}. The dataset contains 188 molecular graphs with an average of approximately 18 nodes and 40 edges per graph. Node attributes consist of categorical atom labels interpreted as node features.

The benchmark is divided into two evaluation settings. In the \textit{ideal graph topology} setting, participants evaluate their models on clean and unmodified node features. In the \textit{perturbed graph topology} setting, node features are corrupted through a combination of feature distribution shift and Gaussian noise injection. Specifically, a constant offset $\delta$ is added to node features to simulate systematic domain shift:
\[
x \leftarrow x + \delta,
\]
where $\delta$ corresponds to a predefined feature shift parameter. In addition, Gaussian noise is injected into node attributes according to:
\[
x \leftarrow x + \epsilon, \qquad \epsilon \sim \mathcal{N}(0,\sigma^2),
\]
where $\sigma$ controls the noise magnitude. This competition falls under the \textcolor{catGen}{generalization} category due to pertubred test graphs.

The dataset is partitioned into training and hidden evaluation splits using stratified sampling. The training split contains labeled molecular graphs together with graph identifiers and binary labels, while the evaluation split provides only graph identifiers and hides the ground-truth labels used for leaderboard scoring. Models are therefore trained exclusively on clean graph features and subsequently evaluated on both clean and perturbed graph representations.

Performance is evaluated using the macro  F1 score, which balances precision and recall across both classes and provides a more reliable evaluation criterion than classification accuracy alone. Each submission is scored separately under the clean and perturbed settings, yielding two metrics: \textbf{$F_1^{\mathrm{ideal}}$} and \textbf{$F_1^{\mathrm{perturbed}}$}. To quantify robustness, the benchmark additionally computes the \textbf{Robustness Gap}:
\[
\mathrm{Robustness\ Gap} = F_1^{\mathrm{ideal}} - F_1^{\mathrm{perturbed}}.
\]
Smaller robustness gaps indicate models that maintain stable performance under feature corruption.

Participants submit encrypted prediction files for both clean and perturbed evaluation settings. An automated evaluation pipeline decrypts submissions securely, computes official metrics against hidden labels, and updates the public leaderboard automatically. Leaderboard ranking prioritizes the highest perturbed macro-$F_1$ score, followed by the smallest robustness gap as a secondary criterion.

\subsection*{C14: BrainAge}

The \textbf{BrainAge} (Brain-Age Prediction Challenge) challenge is a graph-level regression task defined over structural neuroimaging data. The dataset consists of \textcolor{catHomo}{homogenous} brain graphs $\mathcal{G}=\{G_k\}$, where each graph $G_k=(V_k,E_k)$ is constructed from structural and diffusion MRI data from a specific subject's imaging session. Nodes $V_k$ correspond to cortical brain regions defined by the Desikan-Killiany atlas~\cite{desikan2006automated} ($N = 68$), and edges $E_k$ encode structural connectivity derived from diffusion MRI. Given a graph $G_k$, the objective is to learn a mapping $f(G_k)\rightarrow y_k$, where $y_k\in\mathbb{R}$ denotes the age of an individual at the time of image acquisition.

The dataset is constructed from the OASIS-3 longitudinal neuroimaging cohort~\cite{lamontagne2019oasis}, restricted to subjects who remained cognitively normal across all visits. Cortical measurements were obtained from preprocessed FreeSurfer~\cite{fischl2012freesurfer} outputs available through the Neuroimaging Informatics Tools and Resources Clearinghouse (NITRC). Structural connectivity information is sourced from the PIT Bioinformatics Group~\cite{Kerepesi2017}. The final dataset comprises 431 MRI sessions from 288 unique subjects across multiple longitudinal time points.

For every cortical region, two node features are provided: cortical thickness and gray matter volume. Edge weights are defined by diffusion MRI streamline counts between regions. To prevent information leakage arising from repeated longitudinal observations of the same participant, training, validation, and test splits were performed at the subject level using age-stratified sampling.

For evaluation, participants generate age predictions for test MRI sessions and submit them as an encrypted prediction table through a pull-request-based GitHub workflow. The automated GitHub pipeline validates the submission, decrypts the predictions, computes the evaluation score, and updates the leaderboard upon merge. Since this competition involves a single regression target, performance is evaluated on the hidden test split using MAE.

\subsection*{C15: OctoNode Cup}

\textbf{OctoNode Cup} (GitHub Social Network Classification) is a transductive node classification task focused on classifying users in a \textcolor{catHomo}{homogenous} GitHub social network. In this network, represented as a graph $G=(V,E)$, each node corresponds to a GitHub user and edges represent connections between users. The task aims to predict a binary label for each node based on its associated features. The objective is to learn a mapping $f(v_i) \rightarrow y_i$  where $v_i$ denotes a GitHub user and $y_i \in \{0,1\}$ represents the predicted class, with $1$ corresponding to a Web Developer and $0$ corresponding to a Machine Learning Developer.

The dataset is derived from a real GitHub developer network  introduced by \citep{rozemberczki2019multiscale} and represents the platform as a graph. Each node is identified by a unique ID and is associated with a set of features used for prediction. The dataset contains approximately 37,700 nodes and 289,000 edges. It is divided into two partitions: a training set comprising 70\% of the data, which includes both features and labels, and a test set comprising the remaining 30\%, where only the features are provided while the labels remain hidden for evaluation purposes.

Participants are provided with input files required for training and evaluation. Predictions must be generated locally and encrypted prior to submission using the provided encryption method. Only encrypted submission files are uploaded through the repository-based submission system. An automated GitHub pipeline validates the submission, decrypts the predictions, and computes the evaluation score. Submissions are evaluated on the hidden test set using the Macro F1-score.

\subsection*{C16: Graph4ASD Challenge}

The \textbf{Graph4ASD} (Graph Neural Networks for the classification of Autism Spectrum Disorder) challenge is a  binary graph classification task over brain networks derived from resting-state functional magnetic resonance imaging (rs-fMRI) recordings of the human brain. Each sample corresponds to a single subject and is represented as a \textcolor{catHomo}{homogenous} graph $G_k=(A_k,X_k)$, where $A_k\in\mathbb{R}^{n\times n}$ is the weighted adjacency matrix encoding functional connectivity between brain regions and $X_k$ is the node feature matrix. Given a collection $\mathcal{G}=\{G_k\}$, the task is to learn a mapping $f(G_k)\rightarrow y_k\in\{0,1\}$ that assigns each subject to one of two diagnostic classes: Typical Control (0) or Autism Spectrum Disorder (1).

The dataset is drawn from the Autism Brain Imaging Data Exchange (ABIDE)~\cite{graph4ASDdataset}, a multi-site repository of anonymised rs-fMRI scans collected at seventeen international acquisition sites. The full corpus contains 1{,}009 subjects (516 ASD, 51.14\%; 493 controls, 48.86\%), yielding a near-balanced two-class distribution. Each subject's brain volume is parcellated with the Craddock 200 atlas, so every graph has exactly $n=200$ nodes corresponding to predefined regions of interest, with identical node ordering across samples. A challenge-specific subset, preprocessed exclusively for this benchmark, is partitioned into 484 training and 153 evaluation graphs; participants are required to use the data as released, without alternative preprocessing pipelines or auxiliary neuroimaging corpora.

For every subject in the training and evaluation partitions, participants receive a connectivity matrix and a node feature matrix; the training partition is additionally accompanied by subject-level diagnostic labels. Node features are restricted to identity descriptors, so any discriminative signal must be recovered from the connectivity structure over the 200 Craddock regions alone. The training and evaluation partitions are drawn from non-identical site distributions, requiring models to learn site-invariant connectivity representations. 

For evaluation, participants produce a prediction file mapping each test subject to a predicted class. The file is encrypted locally using the provided public key and submitted through a pull-request-based workflow. An automated pipeline validates, decrypts, scores, and updates the leaderboard. Performance on the hidden test split is reported using the macro F1-score.

\subsection*{C17: Mix2Print}

\textbf{Mix2Print} (Learning Material Interaction Physics for identifying parameters of 3D Bioprinting) is a graph regression challenge focused on predicting optimal 3D bioprinting parameters from bioink formulation graphs. The competition models each bioink formulation as a \textcolor{catHomo}{homogenous} graph, where nodes correspond to biomaterials and edges represent potential interactions between materials in the mixture. The task aims to learn the underlying material interaction physics governing extrusion-based bioprinting.

Given a graph formulation $G_i = (V_i, E_i, X_i)$, participants are required to predict three continuous targets: $f(G_i) \rightarrow y_i$ where $y_i = (\text{pressure}, \text{temperature}, \text{speed})$ corresponding to Pressure (kPa), which represents extrusion force; Temperature ($^\circ$C), which represents printing temperature; and Speed (mm/s), which represents print-head velocity.

The dataset represents bioink formulations as fully connected graphs, reflecting the assumption that all biomaterials in a formulation can potentially interact with each other. Node features consist of one-hot encoded material identity features and normalized concentration value. Thus, the total node feature dimensionality is approximately $D \approx 31$. The dataset contains 423 bioink formulations collected from peer-reviewed publications, 30 biomaterials appearing at least five times ,303 training samples and 120 test samples. The train-test split follows a stratified grouping strategy based on printing temperature regimes (hydrogel versus thermoplastic formulations). The dataset captures diverse biomaterial categories including alginates, gelatins, polymers, ceramics, and natural biomaterials.

Participants are provided with graph-based input files for model development and evaluation. Participants are required to generate predictions for the hidden test set in CSV format. To preserve leaderboard integrity, all submissions must be encrypted prior to upload using the provided encryption utility. The encrypted submission file is then uploaded through a GitHub pull request. Since this competition involves three regression targets that may have different value ranges, submissions are evaluated using NMAE, which normalizes the errors across targets before aggregation.

\subsection*{C18: THInC }

\textbf{THInC} (\textbf{T}umor \textbf{H}istopathology \textbf{In}ductive \textbf{C}lassification) is formulated as an \textbf{inductive node classification problem} over \textcolor{catSpatial}{spatial} cell graphs derived from H\&E-stained histology images. Each graph $G_k=(V_k,E_k)$ corresponds to a single patient or tissue slide, where nodes represent individual cell nuclei and edges encode spatial relationships between nearby cells. Given a set of training graphs $\mathcal{G}_{train}$, the goal is to learn a mapping $f(v_i)\rightarrow y_i$ that assigns each cell to one of four classes: Tumor (0), Stromal (1), Lymphocyte (2), or Macrophage (3). Importantly, predictions must generalize to entirely unseen graphs corresponding to new patients, making this an inductive rather than transductive learning setting.

The dataset is derived from the \textbf{NuCLS} breast cancer histopathology dataset \citep{THInC} and is transformed into spatial cell graphs using a biologically grounded pipeline. Starting from raw histology images, cell nuclei are detected and represented as bounding boxes, which are then converted into centroid coordinates $(x,y)$ forming graph nodes. Graph edges are constructed using a $k$-nearest neighbor (k-NN) approach with $k=5$, where each cell is connected to its closest spatial neighbors within the same tissue image.

The inductive split is performed at the image level rather than the node level. Specifically, 80\% of tissue images are used for training, while the remaining 20\% are held out as completely unseen test patients. Node features consist of normalized spatial coordinates $(x,y)$ along with morphological descriptors such as nucleus width and height, standardized to zero mean and unit variance to ensure stable training.

To ensure data privacy, participants are required to encrypt their submission CSV files prior to upload. The evaluation system decrypts and scores predictions in a secure environment, preventing exposure of ground-truth labels from unseen patients during the leaderboard process. Performance is evaluated using accuracy. 

\section{Human Evaluation Details}
\label{app:human_eval}

This appendix provides additional implementation details for the human evaluation track of \textsc{GNN-CB}.

\subsection{Human Implementations}
\label{sec:human_implementations}

All human participants follow a uniform implementation protocol designed to ensure reproducibility, prevent test-set leakage, and isolate model design choices as the primary source of performance variation across submissions. Each competition enforces a fixed wall-clock budget of three hours of CPU time per task, with hyperparameter tuning capped at five training runs per task and a single final submission per benchmark version to discourage test-set overfitting. All random seeds are fixed to \texttt{seed = 25}, training loss curves must be plotted to verify proper convergence and detect overfitting (with early stopping permitted), and the exact source code that produced each submitted prediction file is retained alongside the recorded training time and hardware specifications. 

Human participants never access the hidden test instances, their labels, or any hidden-test performance signal. They receive only the designated training and validation splits, and all model selection and hyperparameter tuning are conducted locally using validation-set performance within the five-run budget. Each submitted solution is evaluated only once through the unified hidden-test evaluation pipeline.

Submitted solutions must be implemented using Graph Neural Network architectures only; convolutional, recurrent, and other non-graph spatial deep-learning models are prohibited, and copy-pasting external solutions is not permitted. LLM assistance is restricted to low-level utility functions and must not expose the model to any task-specific information (e.g., competition name, dataset identity, or feature schema) that could induce data leakage. For each competition, every participant maintains a single documentation file containing the hyperparameter-search strategy used (manual, grid, or random), all configurations attempted within the five-run budget, and the exact prompts and LLM-generated snippets reused in the solution. Together, these records provide a post-hoc audit trail for verifying that the submitted prediction file is consistent with the declared GNN-based solution, documented hyperparameter search, and permitted low-level LLM assistance.


\subsection{Benchmark Composition Attempted by Human Participants}
\label{app:human_composition}

\begin{figure*}[!t]
\centering
\includegraphics[width=0.9\textwidth]{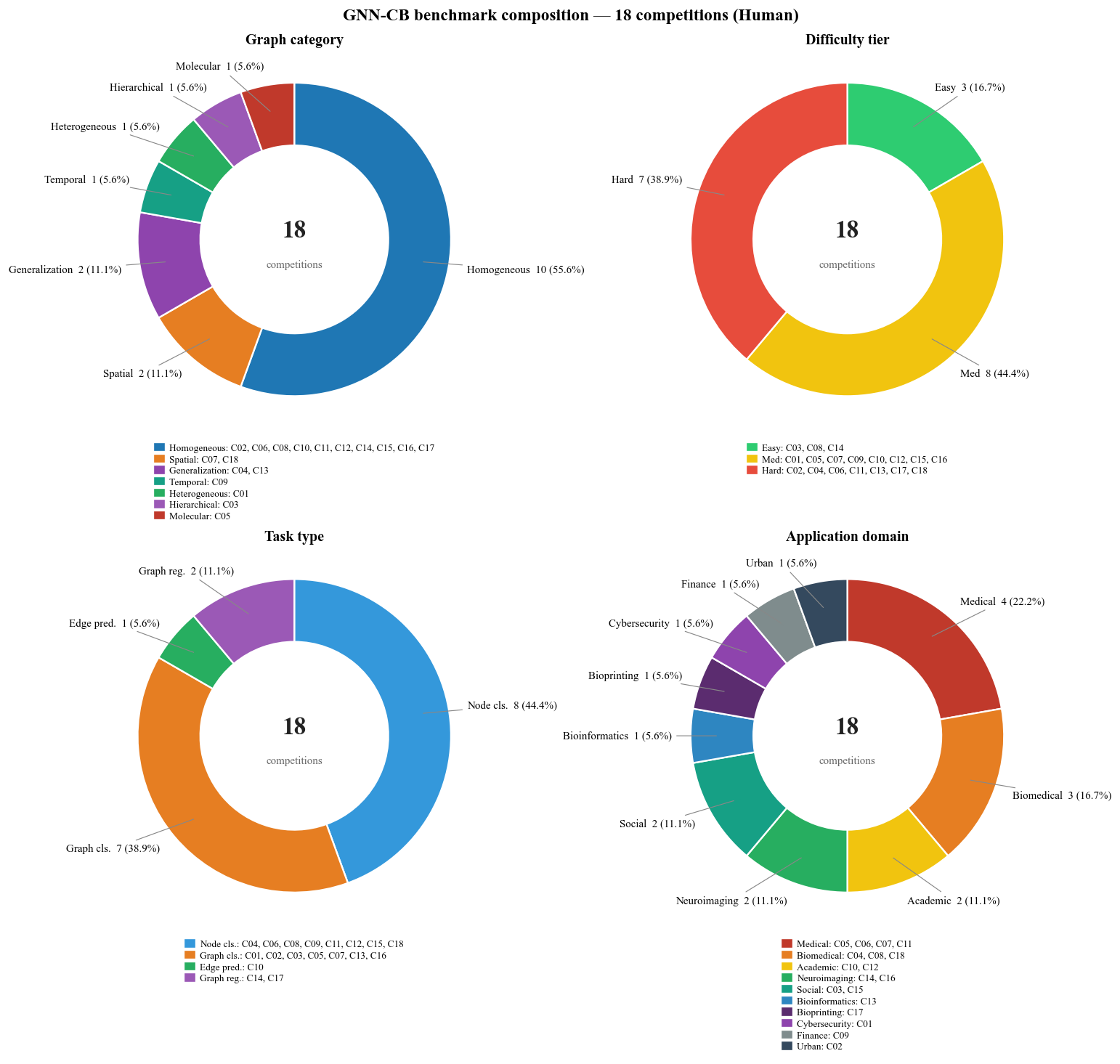}
\caption{Composition of the 18 \textsc{GNN-CB} competitions
attempted by human participants, summarized along four axes:
graph category (top-left), difficulty tier (top-right), task
type (bottom-left), and application domain (bottom-right).
Per-slice counts and competition codes are listed in the legend
below each donut.}
\label{fig:human_composition}
\end{figure*}

Figure~\ref{fig:human_composition} summarizes the composition of
the 18 \textsc{GNN-CB} competitions attempted by human
participants across four characterizing dimensions: graph
category, difficulty tier, task type, and application domain.
The benchmark is dominated by homogeneous graphs ($55.6\%$),
with the remainder distributed across spatial, generalization,
temporal, heterogeneous, hierarchical, and molecular settings.
Difficulty tiers are skewed toward Medium ($44.4\%$) and Hard
($38.9\%$) competitions, with only three Easy tasks. Node classification and graph
classification together account for $83.3\%$ of the tasks,
complemented by one edge-prediction and two graph-regression
competitions. Application domains span medical, biomedical,
academic, neuroimaging, social, bioinformatics, bioprinting,
cybersecurity, finance, and urban data, reflecting the
benchmark's emphasis on real-world heterogeneity rather than a
single application focus.

\subsection{Geographic Distribution of Participants}
\label{app:participant_geography}

\begin{figure*}[!t]
\centering
\includegraphics[width=0.95\textwidth]{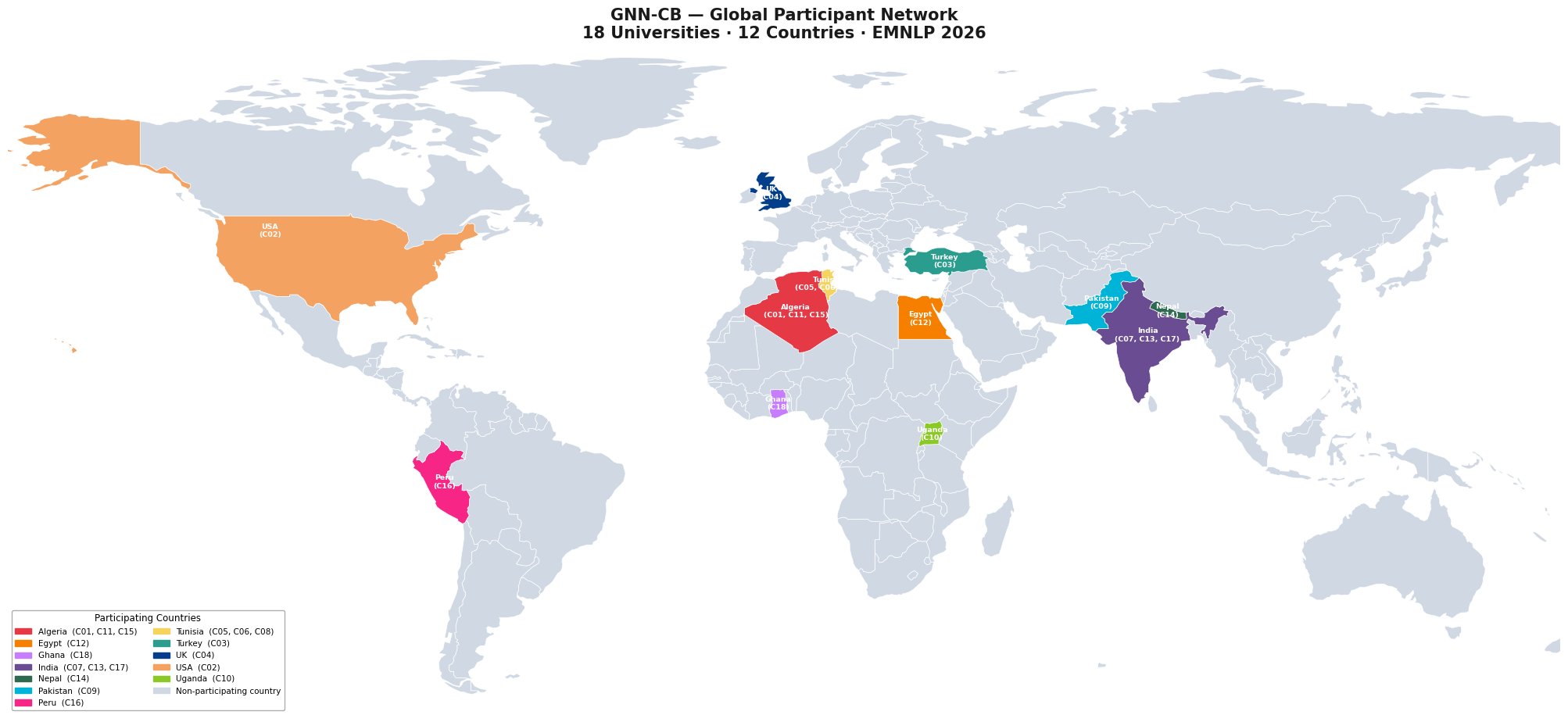}
\caption{Global distribution of the 18 participating
universities that co-designed and evaluated the \textsc{GNN-CB}
benchmark. Each highlighted country is annotated with the
competition(s) contributed by its team(s); the colour key
appears in the legend. The cohort spans 12 countries across
Africa, the Americas, Asia, and Europe.}
\label{fig:participants_world}
\end{figure*}

\textsc{GNN-CB} draws on a globally distributed cohort of human
contributors. Figure~\ref{fig:participants_world} shows the
locations of the 18 participating universities across 12
countries spanning four continents, with each country annotated
by the competition(s) co-designed by its contributing team(s).
The cohort represents Africa (Algeria, Egypt, Ghana, Tunisia,
Uganda), the Americas (Peru, USA), Asia
(India, Nepal, Pakistan), and Europe (Turkey, UK), reflecting
the benchmark's emphasis on broad geographic and institutional
representation rather than concentration in a single research
hub. This geographic diversity is intentional: it ensures that
the curated competitions, datasets, and domain conventions in
\textsc{GNN-CB} are not dominated by the priorities of any
single regional research community, and that the human
performance baseline reported in Table~\ref{tab:human} reflects
a heterogeneous expert pool rather than a homogeneous local
cohort.

\subsection{Human Evaluation Pipeline}
\label{app:human_pipeline}

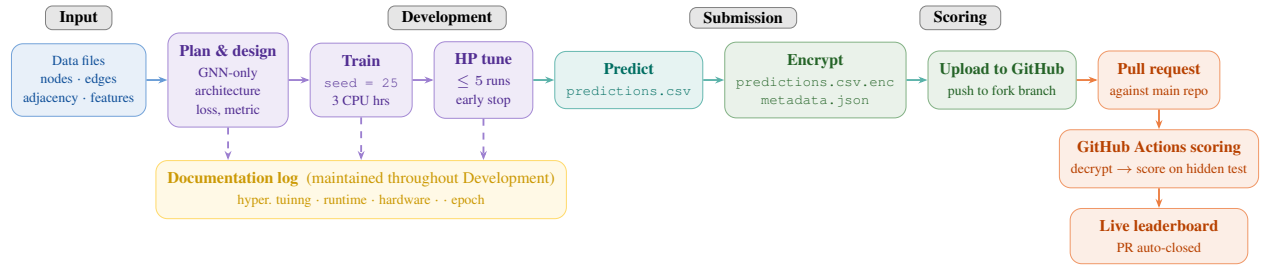
\begin{figure*}[t]
\centering
\resizebox{\textwidth}{!}{%
\begin{tikzpicture}[
  node distance=0.4cm and 0.35cm,
  every node/.style={font=\tiny},
]

\node[ppill=GNNsky]      (ph1) at (0.4,0)    {Input};
\node[ppill=GNNlavender] (ph2) at (6.0,0)    {Development};
\node[ppill=GNNmint]     (ph3) at (10.5,0)   {Submission};
\node[ppill=GNNcoral]    (ph4) at (13.8,0)   {Scoring};



\node[ibox=1.6cm, below=0.2cm of ph1] (input)
  {{\tiny Data files}\\{\tiny nodes $\cdot$ edges}\\{\tiny adjacency $\cdot$ features}};

\node[gbox=1.55cm, right=0.32cm of input] (plan)
  {\textbf{Plan \& design}\\[1pt]
   {\tiny GNN-only}\\{\tiny architecture}\\{\tiny loss, metric}};

\node[gbox=1.55cm, right=0.32cm of plan] (train)
  {\textbf{Train}\\[1pt]
   {\tiny \texttt{seed = 25}}\\{\tiny 3 CPU hrs}};

\node[gbox=1.5cm, right=0.32cm of train] (tune)
  {\textbf{HP tune}\\[1pt]
   {\tiny $\leq 5$ runs}\\{\tiny early stop}};

\node[tbox=1.55cm, right=0.32cm of tune] (preds)
  {\textbf{Predict}\\[1pt]
   {\tiny \texttt{predictions.csv}}};

\node[ebox=1.7cm, right=0.32cm of preds] (encrypt)
  {\textbf{Encrypt}\\[1pt]
   {\tiny \texttt{predictions.csv.enc}}\\{\tiny \texttt{metadata.json}}};

\node[ebox=1.7cm, right=0.32cm of encrypt] (upload)
  {\textbf{Upload to GitHub}\\[1pt]
   {\tiny push to fork branch}};

\node[abox=1.7cm, right=0.32cm of upload] (pr)
  {\textbf{Pull request}\\[1pt]
   {\tiny against main repo}};

\node[abox=2.9cm, below=0.3cm of pr] (score)
  {\textbf{GitHub Actions scoring}\\[1pt]
   {\tiny decrypt $\to$ score on hidden test}};

\node[abox=2.7cm, below=0.3cm of score] (leader)
  {\textbf{Live leaderboard}\\[1pt]
   {\tiny PR auto-closed}};

\node[rbox=5.2cm, below=0.6cm of train, yshift=-0.05cm] (logdoc)
  {\textbf{Documentation log}\enspace(maintained throughout Development)\\[1pt]
   {\tiny hyper. tuinng $\cdot$ runtime $\cdot$ hardware $\cdot$  $\cdot$ epoch}};

\draw[blarr]   (input)   -- (plan);
\draw[purrarr] (plan)    -- (train);
\draw[purrarr] (train)   -- (tune);
\draw[tealarr] (tune)    -- (preds);
\draw[tealarr] (preds)   -- (encrypt);
\draw[tealarr] (encrypt) -- (upload);
\draw[orarr]   (upload)  -- (pr);

\draw[orarr] (pr.south)    -- (score.north);
\draw[orarr] (score.south) -- (leader.north);

\draw[purrarr, dashed] (plan.south)  -- (logdoc.north -| plan.south);
\draw[purrarr, dashed] (train.south) -- (logdoc.north -| train.south);
\draw[purrarr, dashed] (tune.south)  -- (logdoc.north -| tune.south);

\node[font=\scriptsize, text=GNNgray!65,
      above=0.5cm] at ($(input.east)!0.5!(plan.west)$)        {};
\node[font=\scriptsize, text=GNNgray!65,
      above=0.5cm] at ($(preds.east)!0.5!(encrypt.west)$)     {};
\node[font=\scriptsize, text=GNNgray!65,
      above=0.5cm] at ($(encrypt.east)!0.5!(upload.west)$)    {};

\end{tikzpicture}%
}

\caption{Human evaluation pipeline of \textsc{GNN-CB}. Participants prepare graph data (nodes, edges,
adjacency, and feature matrices) from the released data files, then design and train a GNN-only solution under fixed reproducibility constraints (\texttt{seed = 25}, three CPU hours, at most five hyperparameter-tuning runs, early stopping permitted). They generate a \texttt{predictions.csv} file, encrypt it locally into
\texttt{predictions.csv.enc} together with a \texttt{metadata.json},
push the encrypted submission to their GitHub fork, and open a
pull request against the main repository. An automated GitHub
Actions workflow decrypts the submission, scores it against the
hidden test labels, updates the live leaderboard, and closes the
pull request. A documentation log---maintained throughout
development---records the hyperparameter-search strategy, runtime,
hardware specifications, loss-curve verification, and any
LLM-assisted utility usage.}
\label{fig:human_pipeline}

\end{figure*}

Figure~\ref{fig:human_pipeline} summarizes the end-to-end human
evaluation workflow used in \textsc{GNN-CB}, from repository fork
through encrypted pull-request submission to automated GitHub
Actions scoring.

\section{LLM Evaluation Details}
\label{app:llm_eval}

This appendix provides additional implementation details for the LLM evaluation framework used in \textsc{GNN-CB}, including the frozen prompt construction, a representative plan-then-code example, the prompt template, and the orchestration pipeline used during API-based evaluation. As in the human evaluation track, LLMs never access the hidden test set or its labels. Their only pre-submission feedback consists of local execution tracebacks generated during the bounded execute-and-repair loop. Table~\ref{tab:llms} summarizes the set of LLMs evaluated in GNN-CB, grouped by access type (proprietary vs. open-source) and functional category (chat vs. reasoning). For each model, we report the exact API engine identifier used in our orchestration pipeline along with a reference to the official provider documentation.
\begin{table*}[!t]
\centering
\small
\setlength{\tabcolsep}{6pt}
\renewcommand{\arraystretch}{1.15}
\caption{List of LLMs evaluated in \textsc{GNN-CB}, grouped by
access type (proprietary vs.\ open-source). Default decoding parameters are
\texttt{temperature = 0} and \texttt{max\_output\_tokens = 10K},
applied where the provider's API exposes these settings.
API-only (non-agent) runs follow this per-call output budget,
while agent-mode runs may consume substantially more tokens
across multiple internal tool calls and extended-thinking
traces, governed by each agent's native session budget rather
than the per-call cap. The \textbf{Model Engine Name} column
lists the exact API identifier used in our orchestration
scripts; the \textbf{Source} column links to the provider's
official model documentation.}
\label{tab:llms}

\begin{tabular}{@{} c l l c @{}}
\toprule
\textbf{\#} & \textbf{Model Name} & \textbf{Model Engine Name} & \textbf{Source} \\
\midrule

\multicolumn{4}{c}{\textit{Proprietary Chat LLMs}} \\
\midrule
1 & Claude Opus 4.6~\cite{claude_opus_4_6}             & \texttt{claude-opus-4-6}            & \href{https://docs.anthropic.com/en/docs/about-claude/models}{Link} \\
2 & Gemini-3 Flash~\cite{gemini_3_flash}               & \texttt{gemini-3.0-flash}           & \href{https://ai.google.dev/gemini-api/docs/models}{Link} \\
\midrule

\multicolumn{4}{c}{\textit{Open-source Chat LLMs}} \\
\midrule
3 & Llama-3.3 70B~\cite{meta_llama33_70b_instruct}     & \texttt{Llama-3.3-70B-Instruct}     & \href{https://huggingface.co/meta-llama/Llama-3.3-70B-Instruct}{Link} \\
4 & Qwen2.5-Coder 32B~\cite{qwen25_coder_32b_instruct} & \texttt{Qwen2.5-Coder-32B-Instruct} & \href{https://huggingface.co/Qwen/Qwen2.5-Coder-32B-Instruct}{Link} \\
5 & Kimi k2.6~\cite{kimi_k2_6}                         & \texttt{kimi-k2.6}                  & \href{https://platform.moonshot.cn/docs}{Link} \\
\midrule

\multicolumn{4}{c}{\textit{Open-source Reasoning LLMs}} \\
\midrule
6 & DeepSeek v4 Pro~\cite{deepseek_v4_pro}             & \texttt{deepseek-v4-pro}            & \href{https://api-docs.deepseek.com/}{Link} \\
\midrule

\multicolumn{4}{c}{\textit{Proprietary Reasoning LLMs}} \\
\midrule
7 & Gemini 2.5 Pro~\cite{gemini_2_5_pro}               & \texttt{gemini-2.5-pro}             & \href{https://ai.google.dev/gemini-api/docs/models}{Link} \\
8 & Claude Opus 4.7~\cite{claude_opus_4_7}             & \texttt{claude-opus-4-7}            & \href{https://docs.anthropic.com/en/docs/about-claude/models}{Link} \\
9 & GPT-5.4 Codex~\cite{gpt54_codex}                   & \texttt{gpt-5.4-codex}              & \href{https://platform.openai.com/docs/models}{Link} \\

\bottomrule
\end{tabular}
\end{table*}

\subsection{Frozen Prompt Construction}


\begin{figure*}[t]
\centering
\begin{tikzpicture}[
  node distance=0.65cm and 0.7cm,
  every node/.style={font=\tiny},
]

\node[ibox=2.35cm, minimum height=3.0cm, align=left,
      inner sep=4pt] (repo)
{%
  \textbf{Repository}\\[5pt]
  {\tiny\ttfamily README.md}\\[1pt]
  {\tiny\ttfamily /data/}\\[1pt]
  {\tiny\ttfamily tree.txt}\\[1pt]
  {\tiny\ttfamily submit\_spec.md}\\[1pt]
  {\tiny\ttfamily requirements.txt}\\[1pt]
  {\tiny\ttfamily .github/workflow/}%
};

\node[gbox=3.0cm, minimum height=3.0cm, align=center,
      inner sep=3pt, right=0.75cm of repo] (slots)
{%
  \textbf{prompt.md}\\[2pt]
  {\tiny slot filling}\\[2pt]
  \begin{tabular}{c}
    \tikz\node[stag]{\{\{README\}\}};  \\[0.5pt]
    \tikz\node[stag]{\{\{DATA\}\}};    \\[0.5pt]
    \tikz\node[stag]{\{\{TREE\}\}};    \\[0.5pt]
    \tikz\node[stag]{\{\{SUBMIT\}\}};  \\[0.5pt]
    \tikz\node[stag]{\{\{REQS\}\}};
  \end{tabular}%
};

\node[gbox=2.35cm, minimum height=3.0cm, align=center,
      right=0.75cm of slots] (frozen)
{%
  \textbf{Frozen prompt}\\[1pt]
  {\tiny task-agnostic}\\[1pt]
  {\tiny never modified}\\[1pt]
  {\tiny shared across}\\
  {\tiny all 18 competitions}\\[2pt]
  {\tiny\itshape zero-shot}%
};

\draw[GNNpurple!35, dashed, rounded corners=4pt, thin]
  ($(frozen.north west)+(-0.14,0.16)$)
  rectangle
  ($(frozen.south east)+(0.14,-0.16)$);

\node[font=\tiny\itshape, text=GNNpurple!55,
      above=0.2cm of frozen]
  {};

\node[tbox=2.1cm, minimum height=3.0cm, align=center,
      right=0.75cm of frozen] (llmbox)
{%
  \textbf{LLM}\\[6pt]
  {\tiny GPT-5.4}\\[1pt]
  {\tiny Claude\,opus}\\[1pt]
  {\tiny Gemini\,2.5}\\[1pt]
  {\tiny Llama-3-70B}%
};

\node[ebox=2.4cm, minimum height=3.0cm, align=center,
      inner sep=7pt, right=0.75cm of llmbox] (output)
{%
  \textbf{Output}\\[2pt]
  \begin{tabular}{c}
    \tikz\node[draw=GNNteal!55, fill=GNNteal!15,
               text=GNNteal!80!black,
               rounded corners=3pt,
               font=\scriptsize\ttfamily\bfseries,
               minimum width=1.cm,
               minimum height=0.3cm,
               inner xsep=4pt]{$\langle$plan$\rangle$};  \\[1pt]
    \tikz\node[draw=GNNteal!55, fill=GNNteal!15,
               text=GNNteal!80!black,
               rounded corners=3pt,
               font=\scriptsize\ttfamily\bfseries,
               minimum width=1.cm,
               minimum height=0.3cm,
               inner xsep=4pt]{$\langle$code$\rangle$};
  \end{tabular}\\[1pt]
  {\scriptsize standalone}\\
  {\scriptsize executable}%
};

\draw[blarr]   (repo)   --
  node[above, font=\tiny, text=GNNgray!65] {extract}
  (slots);
\draw[purrarr] (slots)  --
  node[above, font=\tiny, text=GNNgray!65] {instantiate}
  (frozen);
\draw[purrarr] (frozen) --
  node[above, font=\tiny, text=GNNgray!65] {prompt}
  (llmbox);
\draw[tealarr] (llmbox) --
  node[above, font=\tiny, text=GNNgray!65] {generate}
  (output);

\end{tikzpicture}

\caption{%
  A frozen prompt construction is instantiated from the repo, and then it is passed to the LLM, which outputs a plan and code.
}
\label{fig:prompt_instantiation}
\end{figure*}
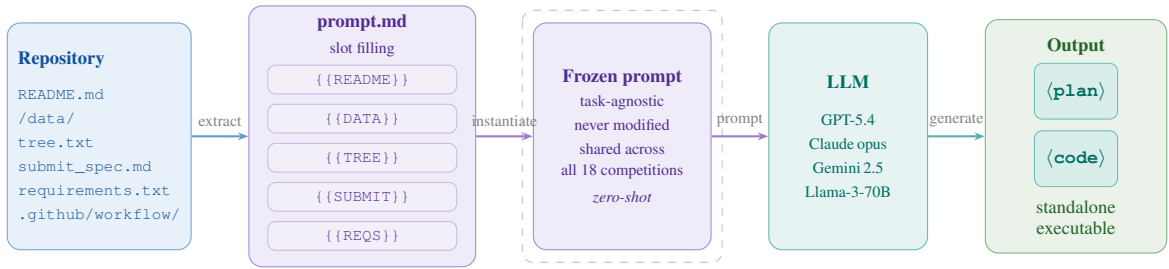

Figure~\ref{fig:prompt_instantiation} illustrates how repository components are inserted into structured slots without manual modification, producing an instantiated prompt shared across all competitions.

\subsection{Plan-Then-Code Example}


\begin{figure*}[t]
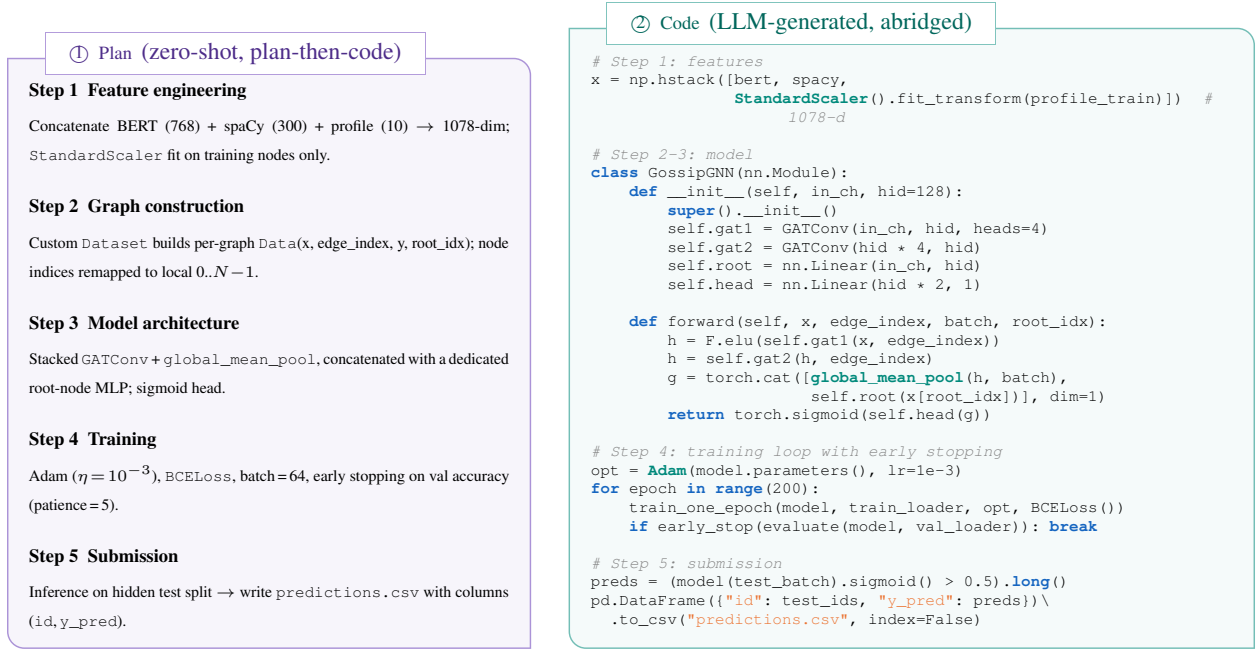

\centering
\begin{minipage}[t]{0.42\linewidth}
\begin{tcolorbox}
[plansty, title={\normalfont\textcircled{\scriptsize 1}\enspace Plan\enspace{\small (zero-shot, plan-then-code)}}]
\setlength{\parskip}{2.5pt}

{\scriptsize\bfseries Step 1\enspace Feature engineering}\par
{\tiny Concatenate BERT (768) + spaCy (300) + profile (10) $\to$ 1078-dim;
       \texttt{StandardScaler} fit on training nodes only.}

\medskip
{\scriptsize\bfseries Step 2\enspace Graph construction}\par
{\tiny Custom \texttt{Dataset} builds per-graph \texttt{Data}(x, edge\_index, y, root\_idx);
       node indices remapped to local 0..$N{-}1$.}

\medskip
{\scriptsize\bfseries Step 3\enspace Model architecture}\par
{\tiny Stacked \texttt{GATConv} + \texttt{global\_mean\_pool}, concatenated
       with a dedicated root-node MLP; sigmoid head.}

\medskip
{\scriptsize\bfseries Step 4\enspace Training}\par
{\tiny Adam ($\eta\!=\!10^{-3}$), \texttt{BCELoss}, batch\,=\,64,
       early stopping on val accuracy (patience\,=\,5).}

\medskip
{\scriptsize\bfseries Step 5\enspace Submission}\par
{\tiny Inference on hidden test split $\to$ write
       \texttt{predictions.csv} with columns (\texttt{id},\,\texttt{y\_pred}).}

\end{tcolorbox}
\end{minipage}%
\hfill%
\begin{minipage}[t]{0.55\linewidth}
\begin{tcolorbox}[codesty, title={\normalfont\textcircled{\scriptsize 2}\enspace Code\enspace{\small (LLM-generated, abridged)}}]

\begin{lstlisting}[style=GNNcode]
# Step 1: features
x = np.hstack([bert, spacy,
               StandardScaler().fit_transform(profile_train)])  # 1078-d

# Step 2-3: model
class GossipGNN(nn.Module):
    def __init__(self, in_ch, hid=128):
        super().__init__()
        self.gat1 = GATConv(in_ch, hid, heads=4)
        self.gat2 = GATConv(hid * 4, hid)
        self.root = nn.Linear(in_ch, hid)
        self.head = nn.Linear(hid * 2, 1)

    def forward(self, x, edge_index, batch, root_idx):
        h = F.elu(self.gat1(x, edge_index))
        h = self.gat2(h, edge_index)
        g = torch.cat([global_mean_pool(h, batch),
                       self.root(x[root_idx])], dim=1)
        return torch.sigmoid(self.head(g))

# Step 4: training loop with early stopping
opt = Adam(model.parameters(), lr=1e-3)
for epoch in range(200):
    train_one_epoch(model, train_loader, opt, BCELoss())
    if early_stop(evaluate(model, val_loader)): break

# Step 5: submission
preds = (model(test_batch).sigmoid() > 0.5).long()
pd.DataFrame({"id": test_ids, "y_pred": preds})\
  .to_csv("predictions.csv", index=False)
\end{lstlisting}

\end{tcolorbox}
\end{minipage}

\caption{Plan-then-code paradigm illustrated for C03 (Real or Fake!): a 5-step structured plan (left) and the LLM-generated, executable implementation skeleton (right).}
\label{fig:plan_block}
\end{figure*}

Figure~\ref{fig:plan_block} presents a representative plan-then-code output for C03 (Real or Fake!), with the high-level reasoning plan on the left and the LLM-generated executable code on the right.

\subsection{Frozen Prompt Template}

To minimize prompt-engineering bias and ensure cross-model comparability, every LLM evaluation in \textsc{GNN-CB} uses the same task-frozen prompt template, instantiated through automatic slot filling from the competition repository. The template encodes the inputs the model receives, the structured output it must produce, and the execution constraints under which the generated solution will be evaluated. Listing \ref{lst:frozen_prompt} reproduces a simplified version of this template; the same string is shared verbatim across all 18 competitions, with only the slot contents (README, repository tree, dataset summary, submission specification, and allowed libraries) varying per task.

\subsection{LLM Evaluation Pipeline Listings}

The conceptual distinction between the non-agent (API-only) and
agent-based evaluation modes is described in
Section~\ref{subsec:agent_vs_nonagent}. Listing~\ref{lst:nonagent_loop}
reproduces the simplified Python orchestration loop used in the
non-agent mode; the agent-based mode has no equivalent
participant-side script, as it relies on the autonomous coding
system's native repository inspection, code execution, and repair
capabilities, governed by each agent's session budget.

\subsection{Performance Analysis of LLMs Across Competitions}
\label{app:llm-perf}

To assess the comparative performance of leading LLMs against human baselines, we evaluate their capabilities across Easy, Medium, and Hard classification tasks in the \textsc{GNN-CB} benchmark from three perspectives: a gap analysis quantifying performance differences relative to the human-top score, an average human--LLM comparison, and a category-based comparison of how different model types perform across difficulty levels.

\begin{figure*}[!t]
  \centering
  \includegraphics[width=\textwidth]{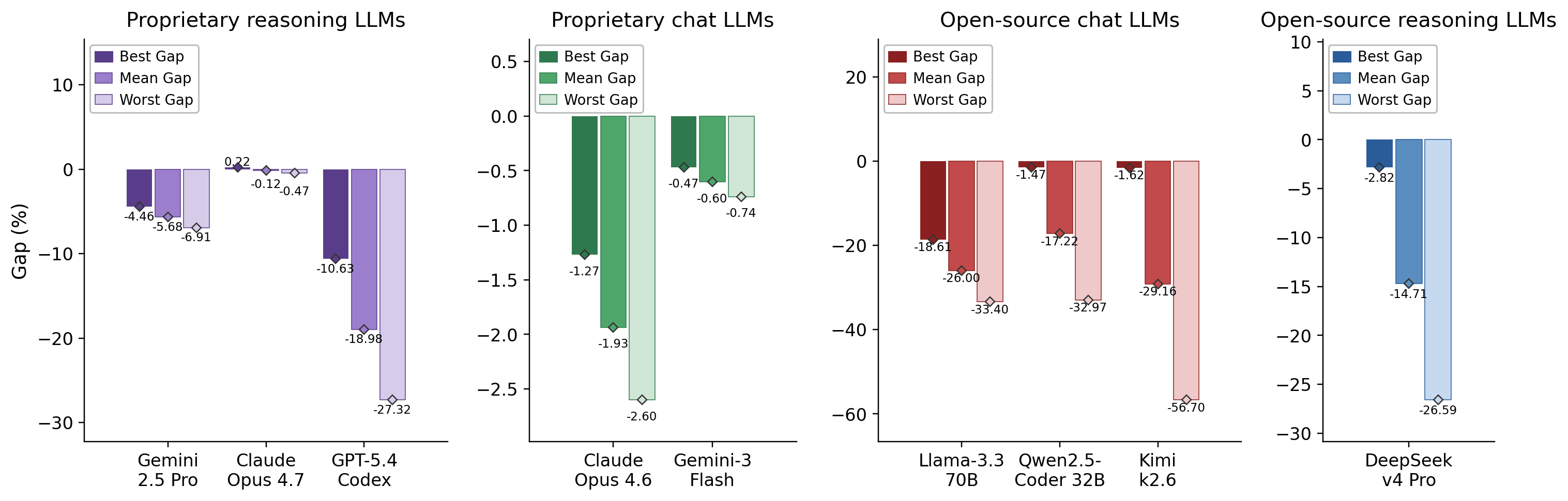}
\caption{Best, Mean, and Worst Gap performance for leading LLMs on the \textbf{Easy} classification tasks of the \textsc{GNN-CB} benchmark, grouped by model family, where the Gap is defined as Gap = Score - Human Top}
  \label{fig:llm-easy}
\end{figure*}

\begin{figure*}[!t]
  \centering
  \includegraphics[width=\textwidth]{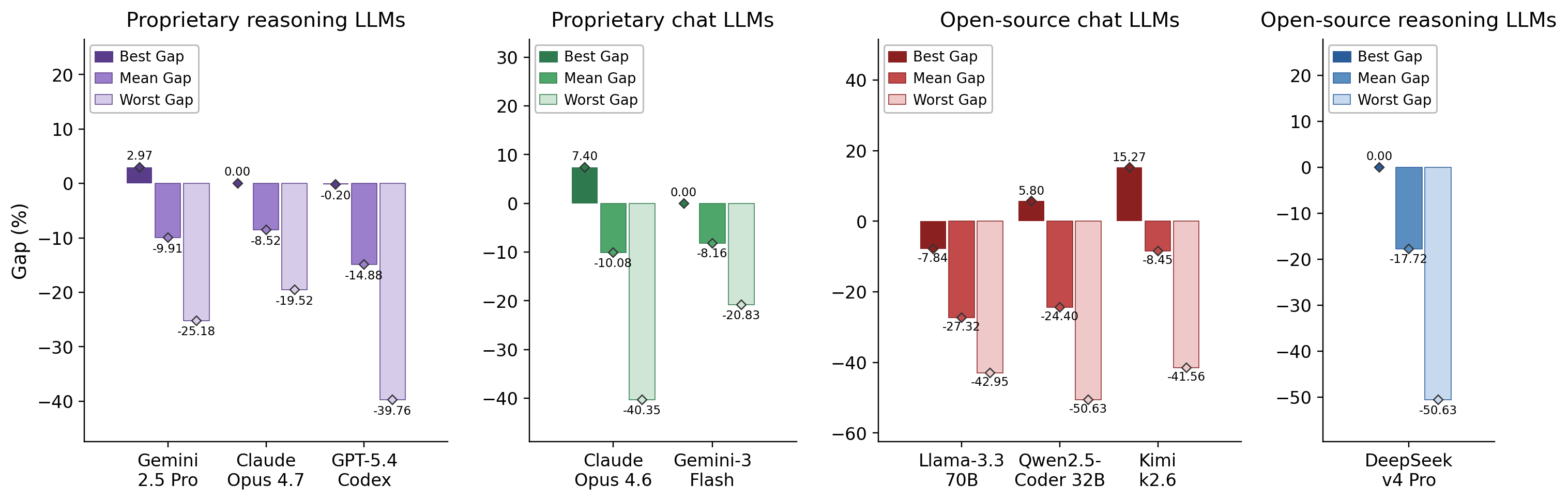}
\caption{Best, Mean, and Worst Gap performance for leading LLMs on the \textbf{Medium} classification tasks of the \textsc{GNN-CB} benchmark, grouped by model family, where the Gap is defined as Gap = Score - Human Top}
  \label{fig:llm-med}
\end{figure*}

\paragraph{Gap analysis:}
For Easy classification tasks (Figure~\ref{fig:llm-easy}), proprietary models except from GPT 5.4 Codex show minimal deviation, with best-case gaps approaching zero and worst-case gaps remaining moderate. In contrast, open-source models exhibit larger negative gaps across all models, indicating instability and weaker alignment with human-top performance. On Medium tasks (Figure~\ref{fig:llm-med}), the performance gap widens across all model categories. Proprietary models continue to outperform open-source models in terms of mean gap and exhibit more consistent behavior overall, whereas open-source models show substantially larger negative gaps and higher variability across competitions.

\begin{lstlisting}[
  language={},
  caption={Simplified frozen prompt template used across all 18 \textsc{GNN-CB} competitions.},
  label={lst:frozen_prompt},
  basicstyle=\scriptsize\ttfamily,
  numbers=none,
  frame=single,
  framesep=4pt,
  xleftmargin=4pt,
  xrightmargin=4pt,
  breaklines=true,
  showstringspaces=false]
You are solving a GNN coding competition.

You will receive:
  (1) competition README
  (2) repository tree
  (3) dataset summary
  (4) submission format
  (5) allowed libraries

Produce:

<plan>
  ...structured reasoning plan...
</plan>

<code>
  ...standalone executable implementation...
</code>

Constraints:
  - runnable as `python solution.py`
  - CPU-only execution
  - no external downloads
  - must generate predictions.csv
\end{lstlisting}

\begin{figure*}[!t]
  \centering
  \includegraphics[width=\textwidth]{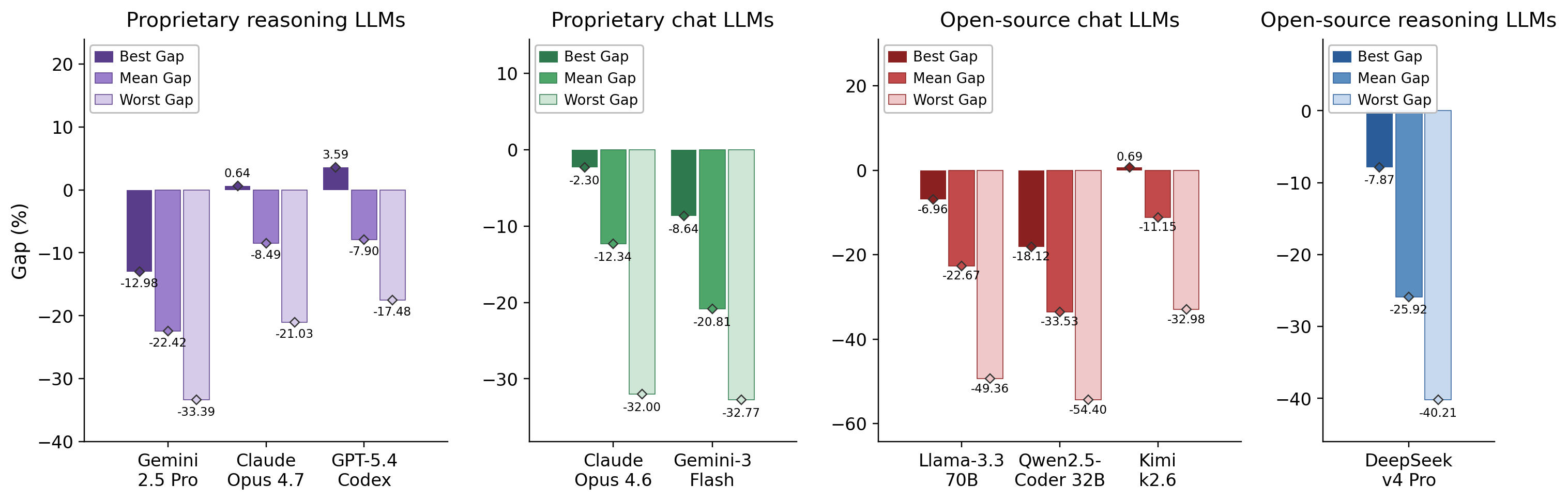}
\caption{Best, Mean, and Worst Gap performance for leading LLMs on the \textbf{Hard} classification tasks of the \textsc{GNN-CB} benchmark, grouped by model family, where the Gap is defined as Gap = Score - Human Top}
  \label{fig:llm-hard}
\end{figure*}

On Hard tasks (Figure~\ref{fig:llm-hard}), the same overall trend persists, although the gap further increases for nearly all models. Notably, the performance of Gemini 2.5 Pro drops considerably, while Kimi k2.6 reaches a comparable level on several tasks. Despite these improvements, the overall gap between LLMs and human participants remains large on hard competitions. Across all difficulty levels, no substantial performance difference is observed between chat and reasoning LLMs.

\begin{lstlisting}[
  language=Python,
  caption={Simplified non-agent orchestration loop with bounded
           execute-and-repair iterations ($k\!\leq\!5$).},
  label={lst:nonagent_loop},
  basicstyle=\footnotesize\ttfamily,
  numbers=none,
  frame=single,
  framesep=4pt,
  xleftmargin=4pt,
  xrightmargin=4pt,
  breaklines=true,
  showstringspaces=false]
prompt = instantiate_frozen_prompt(repo)
for attempt in range(REPAIR_LIMIT):           # k <= 5
    response = call_llm(prompt)               # API request
    plan     = extract_plan(response)
    code     = extract_code(response)
    write_solution(code)                      # -> solution.py
    result   = execute_solution()             # CPU-only sandbox
    if valid_submission(result):              # schema + content
        break
    prompt += format_traceback(result)        # appended for repair
\end{lstlisting}

\begin{figure*}[!t]
  \centering
  \includegraphics[width=\textwidth]{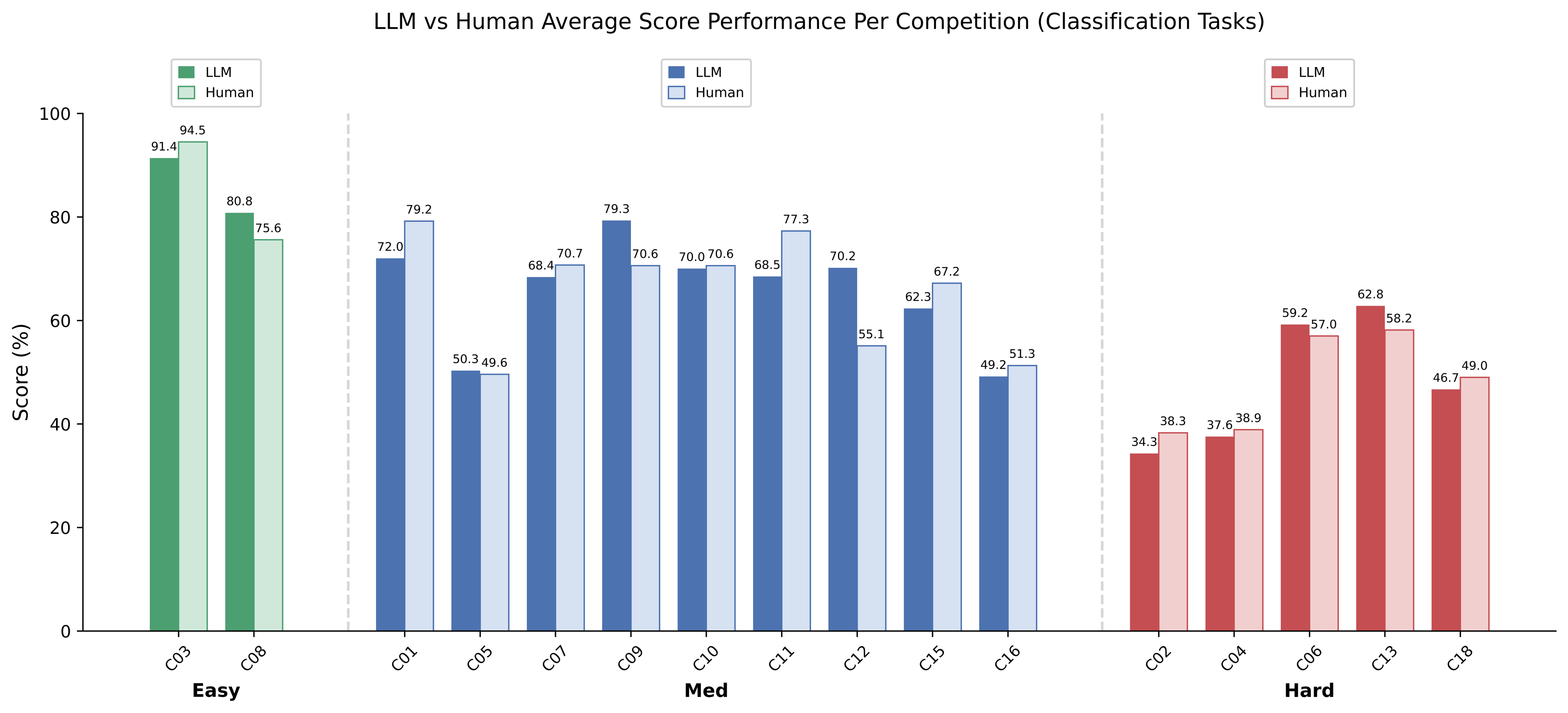}
\caption{LLM vs Human Average Performance across difficulty levels}
  \label{fig:avg-all}
\end{figure*}

\paragraph{Average human--LLM comparison:}
Figure~\ref{fig:avg-all} analyzes LLM and human average scores across Easy, Medium, and Hard competitions on a per-competition basis. Although LLM averages occasionally surpass human averages on individual competitions, humans still achieve higher average scores in 10 out of the 16 classification competitions, indicating that human participants remain more consistently reliable overall. Moreover, as task difficulty increases from Easy to Hard, LLM performance declines similarly to human performance, with larger drops observed on harder tasks.

\begin{figure*}[!t]
  \centering
  \includegraphics[width=\textwidth]{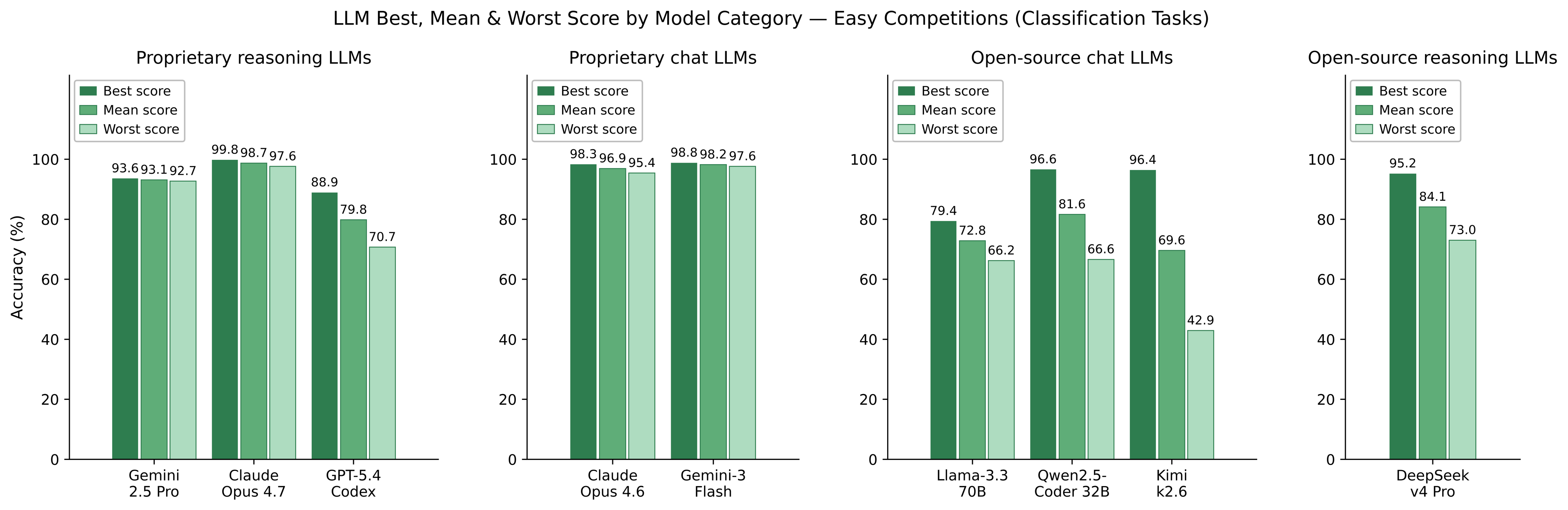}
\caption{LLM performance grouped by model family on \textbf{Easy} classification tasks. Performance metrics capture the best, mean, and worst scores achieved by each model.}
  \label{fig:easy}
\end{figure*}

\paragraph{Category-based comparison:}
Finally, we analyse model performance by category across the Easy, Medium, and Hard classification tasks using each model's best, mean, and worst accuracy. On Easy tasks (Figure~\ref{fig:easy}), proprietary models achieve high performance with a small variablity between best and worst scores, indicating strong consistency, while open-source models remain competitive but display wider variability. On Medium and Hard tasks (Figures~\ref{fig:med} and~\ref{fig:hard}), mean scores decline across nearly all models and the best--worst variance widens, indicating reduced stability as complexity increases. The steep decline in mean and worst-case accuracy highlights the limitations of current foundational architectures under higher complexity.

\clearpage
\onecolumn

\begin{center}
  \includegraphics[width=\textwidth]{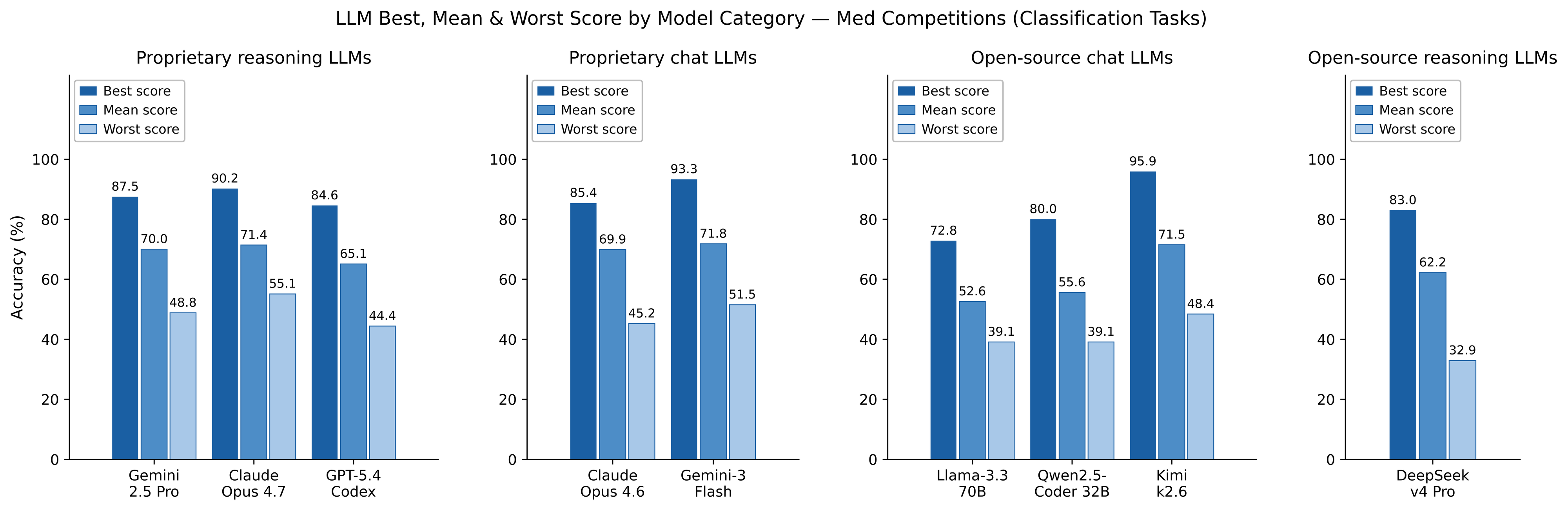}
  \captionof{figure}{LLM performance grouped by model family on \textbf{Medium} classification tasks.}
  \label{fig:med}
\end{center}

\vspace{1em}

\begin{center}
  \includegraphics[width=\textwidth]{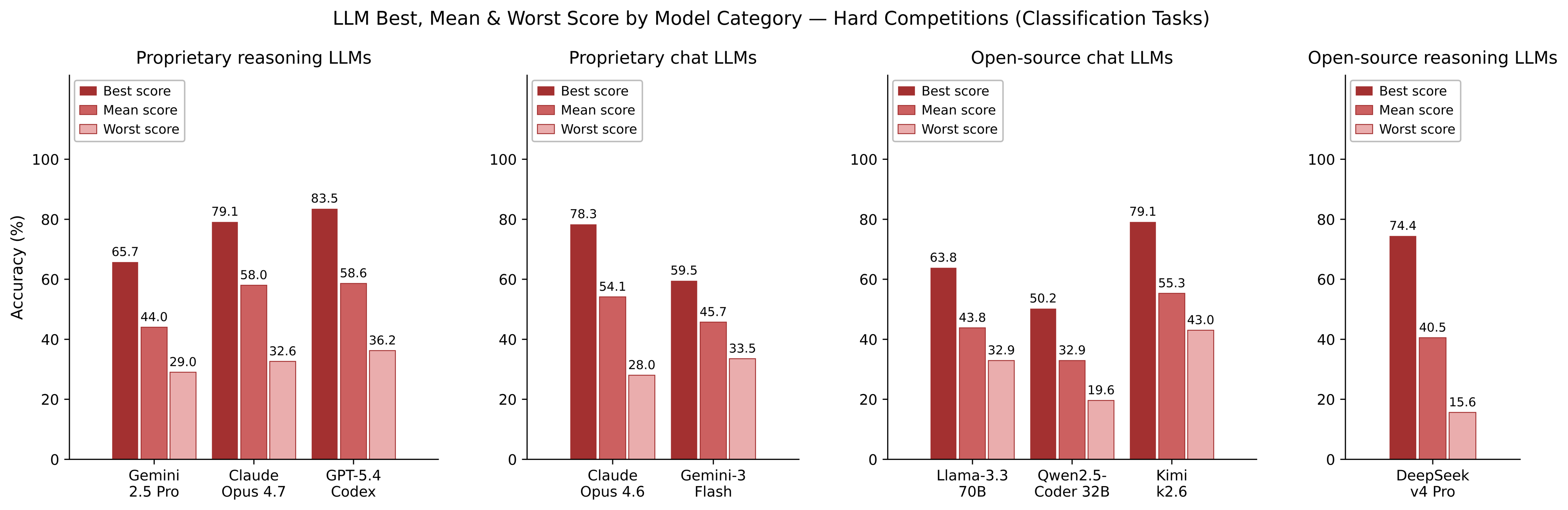}
  \captionof{figure}{LLM performance grouped by model family on \textbf{Hard} classification tasks.}
  \label{fig:hard}
\end{center}

\paragraph{Error Analysis:}
Severe underperformance was concentrated primarily among open-source models: 17 of the 20 cases (85\%) were produced by open-source LLMs, compared with only three (15\%) by proprietary models. When normalized by the number of evaluated model--task pairs, severe failures occurred in 17/64 (26.6\%) open-source pairs and 3/80 (3.8\%) proprietary pairs. Chat models accounted for 14 cases and reasoning models for six, corresponding to failure rates of 14/80 (17.5\%) and 6/64 (9.4\%), respectively. However, given the small and unbalanced number of models across categories, particularly the presence of only one open-source reasoning model, these differences should be interpreted as descriptive patterns rather than inherent effects of model family or accessibility.

\section*{Acknowledgments}

This research was carried out as part of the \href{https://basira-lab.com/}{BASIRA Lab} \textbf{Global Consortia Projects} within the \href{https://www.linkedin.com/feed/update/urn:li:activity:7368982974270242816/}{GNNs for
Rising Stars Program}{}, which Dr Rekik founded, organized, and taught, and
which gave the authors their shared foundation in graph
representation learning and directly shaped the design of the
competitions presented here; further details on the program and its
competitions are available at
\url{https://basiralab.github.io/GNN-CB/}. We are grateful to have carried out this work as a global consortium of early-career researchers spanning 24 institutions across 16 countries.


\end{document}